\pdfoutput=1

\documentclass[11pt]{article}

\usepackage[preprint]{acl}
\usepackage{multirow}
\usepackage{times}
\usepackage{latexsym}

\usepackage[T1]{fontenc}

\usepackage[utf8]{inputenc}

\usepackage{microtype}

\usepackage{inconsolata}

\usepackage{graphicx}

\usepackage{booktabs}
 \usepackage{amsmath} 
 \usepackage{amssymb}
\title{FocusLLM: Precise Understanding of Long Context by \\ Dynamic Condensing}

\author{
 \textbf{Zhenyu Li},
 \textbf{Yike Zhang},
 \textbf{Tengyu Pan},
 \textbf{Yutao Sun},\\
 \textbf{Zhichao Duan},
 \textbf{Junjie Fang},
 \textbf{Rong Han},
 \textbf{Zixuan Wang},
 \textbf{Jianyong Wang\thanks{Corresponding author}}
\\
\\
 \textsuperscript{}Tsinghua University
}

\begin{document}
\maketitle
\begin{abstract}

Empowering LLMs with the ability to precisely understand long contexts is crucial for many downstream applications. However, handling long contexts with conventional transformer architecture requires substantial training and inference resources. Existing context condensing methods cannot accurately understand the full context, as there is a considerable amount of information loss in the condensing process.
To address these issues, we present \textbf{FocusLLM}, a framework designed to extend the fixed context length of any decoder-only LLM, allowing the model to focus on relevant information from very long sequences. 
FocusLLM first divides long text input into chunks based on the model's original context length. It then employs the \textbf{\textit{dynamic condensing}} process to distill crucial information from each chunk. Ultimately, through the novel \textbf{\textit{parallel decoding}} mechanism, FocusLLM can integrate the extracted information into its local context. 
FocusLLM stands out for great training efficiency and versatility: trained with an 8K input length and with much less training cost than previous methods, FocusLLM exhibits superior performance across downstream tasks and maintains strong language modeling ability when handling extensive long texts, even up to 400K tokens. Our code is available at \url{https://github.com/leezythu/focusllm}.

\end{abstract}

\section{Introduction}
The importance of extending the context length of large language models (LLMs) cannot be overstated. In numerous applications, ranging from complex document analysis to generating coherent long-form text, the ability to effectively utilize extended context is critical. For instance, in tasks such as document summarization and question answering over lengthy articles, a more extensive context allows for a more comprehensive understanding and accurate responses. However, leveraging long contexts in LLMs presents several formidable challenges. (1) The computational complexity of transformers \cite{vaswani2017attention} grows quadratically with the sequence length, rendering the training process prohibitively expensive. (2) LLMs exhibit poor extrapolation performance for longer sequences, even after additional fine-tuning \cite{chen2023extending,peng2023yarn}. (3) Acquiring high-quality long-text datasets, which are essential for training and fine-tuning, is exceedingly difficult \cite{xiong2023effective,wang2022self}.
\begin{figure}[t]
  \includegraphics[width=0.9\linewidth]{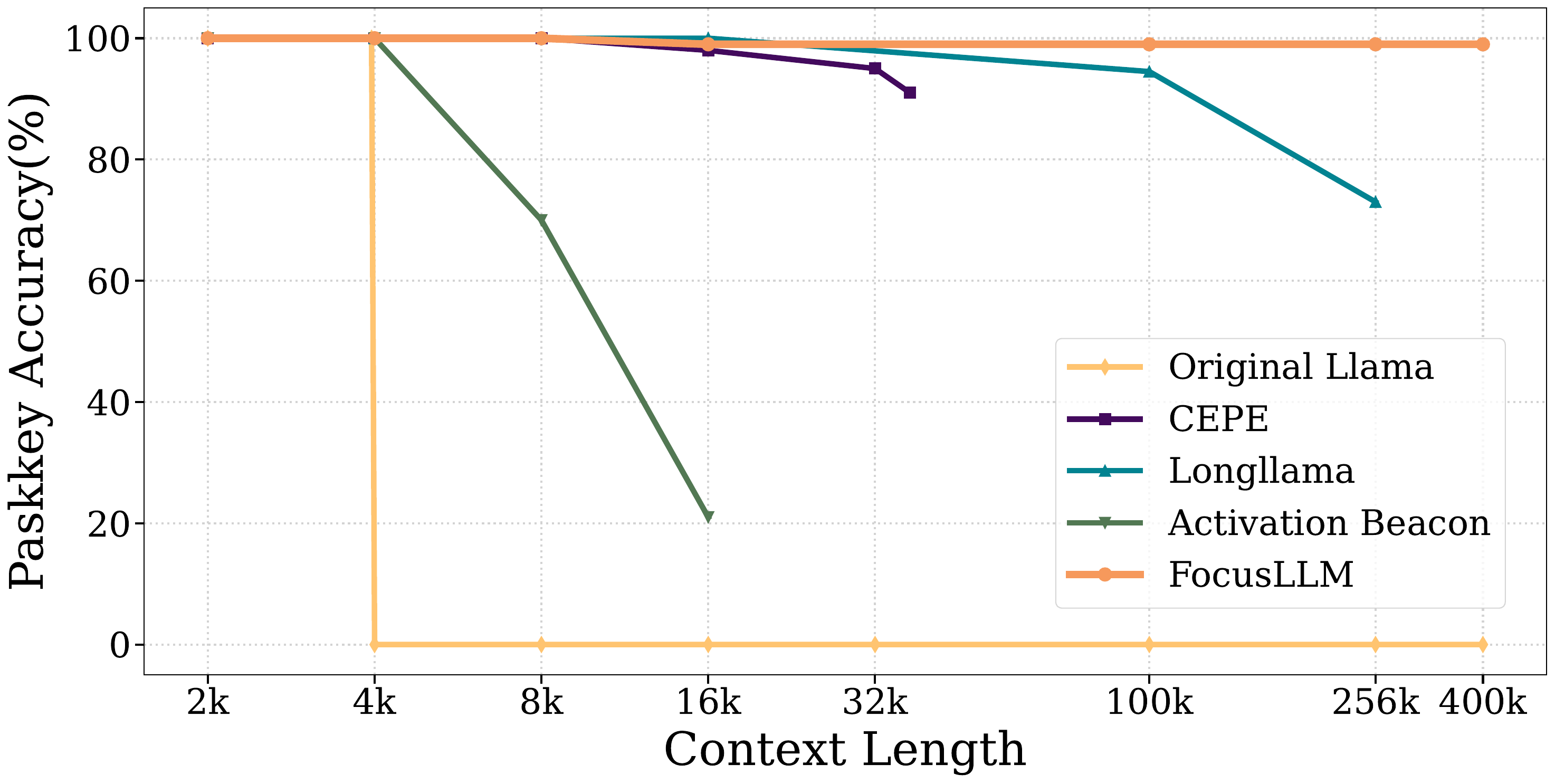}
  \caption{A comparison between FocusLLM and previous context scaling methods on the passkey retrieval task, including CEPE, LongLLaMA and Activation Beacon. Our method extrapolates beyond the original context length of LLaMA, achieving 99\% accuracy at a context length of 400K, with less training cost.}
  \label{fig:passkey}
\end{figure}

To circumvent the substantial costs of directly scaling the window length by continual training on longer inputs, recent work has proposed to drop unimportant tokens and retain important tokens, either by modifying the attention mechanism \cite{xiao2023efficient,han2023lm}  or by compressing the context into some specialized tokens \cite{zhang2024soaring,chevalier2023adapting,ge2023context}, in order to effectively condense long textual information. However, these methods overlook the fact that \textit{token importance changes dynamically during the decoding process}: tokens previously considered unimportant may become crucial in later decoding steps. As a result, they share a common drawback, which we refer to as \textbf{\textit{information loss}}: some tokens that will be needed in the future have already been discarded. For example, in Passkey Retrieval task \cite{mohtashami2024random} illustrated in Figure \ref{fig:passkey}, as the context length increases, the compression method Activation Beacon fails to retrieve passkey pairs that appeared in the earlier context.


\begin{figure*}[htbp]
  \includegraphics[width=\linewidth]{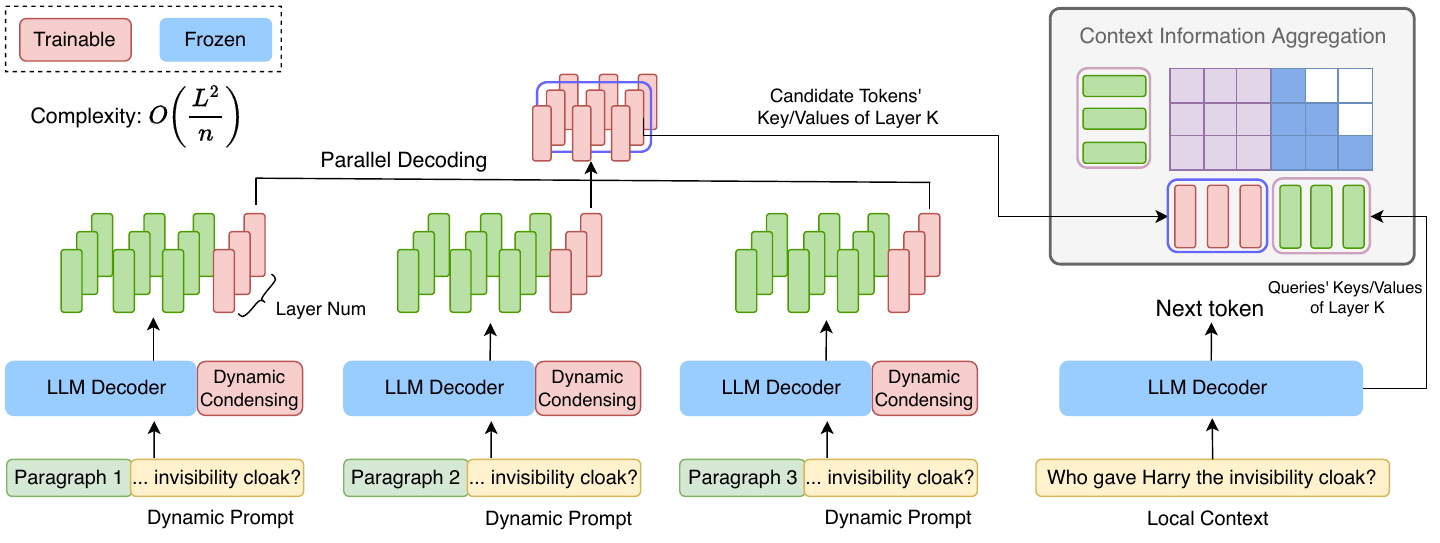}
  \caption{One decoding step of the FocusLLM framework. A small fragment of the local context (denoted as the dynamic prompt) is appended to each chunk. The representations of the candidate tokens, obtained through dynamic condensing and parallel decoding, are then concatenated and integrated back into the local context.}
  \label{fig:framework}
\end{figure*}

Considering the above issues, the question arises: \textit{can we extend the context length of an existing LLM at a low cost without any information loss?} In this paper, we propose a training efficient and effective solution \textbf{\textit{FocusLLM}}, which can maintain a precise understanding of the whole long context. Specifically, FocusLLM first divides a long text into chunks based on the model's original context length. Then, the \textbf{\textit{dynamic condensing}} process is applied, which appends dynamic prompts to each chunk to extract crucial information, ensuring no information loss. Finally, we use \textbf{\textit{parallel decoding}} mechanism to aggregate information from different chunks and generate the next token. The original model parameters are kept frozen to maintain generalization capabilities, with only a small number of trainable parameters introduced for dynamic condensing.

We employ the FocusLLM framework to the widely used LLaMA-2-7B model \cite{touvron2023llama2}, which has a default context length of 4K. In terms of efficiency, FocusLLM is trained on sequences shorter than 8K tokens and only requires a training budget of \textbf{\textit{0.5B tokens}}. To validate the effectiveness of FocusLLM, we evaluate it across a variety of tasks. Initially, we assessed FocusLLM's language modeling capability. FocusLLM maintains low perplexity on documents comprising 128K tokens and even longer sequences.
Subsequently, to comprehensively evaluate the applicability of FocusLLM in real-world scenarios, we utilized two widely used benchmarks: Longbench \cite{bai2023longbench} and $\infty$-Bench \cite{zhang2024inftybench}. Experimental results demonstrate that FocusLLM has achieved superior performance on both benchmarks, surpassing all baselines including length extrapolation models, continual training models, and similar models designed for extreme long sequences. The main contributions of this paper can be summarized as follows:
\begin{itemize}
\item We propose the FocusLLM framework, which leverages novel \textbf{\textit{dynamic condensing}} and \textbf{\textit{parallel decoding}} mechanisms to avoid information loss and achieve precise understanding of long contexts, as shown in Figure \ref{fig:passkey}.

\item Compared to previous context-scaling methods, FocusLLM achieves remarkable results with \textbf{\textit{high training efficiency}} by introducing only a small set of trainable parameters and utilizing a training budget of 0.5B tokens.

\item Through comprehensive evaluation, FocusLLM outperforms all baselines on downstream tasks while maintaining low perplexity, demonstrating that it can seamlessly serve as a general-purpose language model.
\end{itemize}

\section{Architecture}\label{architecture}

The overall framework of FocusLLM is presented in Figure \ref{fig:framework}. Each decoder in the figure shares the same model (e.g. LLaMA-2). 

\subsection{\textbf{Notations}}\label{notations} 
 Given a long sequence with $S$ tokens  $\{x_1,...,x_S\}$, we segment them into \textbf{memory tokens} $\{x_1,...,x_m\}$ and \textbf{local context} $\{x_{m+1},...,x_S\}$, with the length of local context not exceeding the model's default context length, denoted as \( L \). Concurrently, we divide the memory tokens into \textbf{chunks}, labeled as \( C_1, C_2, ..., C_k \), with each chunk's size also not exceeding \( L \). These chunks can represent distinct documents or a single long document. We define the original decoder model as \( F_{\text{dec}} \) and its hidden dimension \( d_{dec} \). To endow the model with the capability for dynamic condensing, we introduce a small set of new parameters, resulting in the modified model \( F'_{\text{dec}} \). 

\subsection{\textbf{Dynamic Condensing}}
As highlighted in the introduction, the importance of tokens in the context dynamically changes at each decoding step. Previous work that condenses context using a fixed pattern suffers from the drawback of \textit{information loss}. To address this issue, we propose the dynamic condensing mechanism, which consists of two key steps: dynamic prompt injection and candidate token generation.
\newline
\textbf{Dynamic Prompt Injection.}
We append a small fragment of local context (we refer to it as the \textit{dynamic prompt} in Figure \ref{fig:framework}) behind each chunk. The motivation is to aggregate the most critical information from each chunk for the current decoding step.
We can formally define this process as follows:
\begin{equation}
\small
     \hat{C_i} \leftarrow \{C_i;x_{m+j},...,x_S\} \ i=1,...,k ; 1 \leq j \leq S-m
\end{equation}
Here \( j \) is a hyperparameter that determines the number of local tokens appended to each chunk.
We adopt a default length of 512 tokens for inference, which is sufficient to encapsulate the necessary local contextual information.

The last token of the dynamic prompt is used to generate candidate tokens, which we will explain in detail later. After each decoding step, when FocusLLM generates the next token, this token will be appended to the dynamic prompt \footnote{The first token of the dynamic prompt can be dropped to maintain its fixed length.}. This updated dynamic prompt is then used to generate new candidate tokens in the next decoding step. 

The \textit{dynamic prompt} evolves with each decoding step, ensuring that the model always has access to the most relevant information for the current step.
\newline
\textbf{Candidate Token Generation.}
Building on the dynamic prompt injection described above, we introduce \textit{candidate tokens} to condense the information from each chunk that is crucial for the current decoding step.
The \textit{candidate token} is denoted as the trainable hidden states corresponding to the last local token \( x_S \) in each chunk $\hat{C_i}$. To obtain the representations of candidate tokens, motivated by \cite{zhang2024soaring}, we add a new set of trainable parameters to the linear projection matrices of each layer, while keeping the original model parameters frozen to preserve its original decoding ability. Formally, the trainable parameters for dynamic condensing are:
\begin{equation}
\small
   \{W^c_Q, W^c_K, W^c_V, W^c_O\}_l
\end{equation}
 where \( W^c_Q \), \( W^c_K \), \( W^c_V \), and \( W^c_O \) represent the new linear projections for the query, key, value, and output matrices associated with the candidate token, and \( l \) denotes the layer number. 
 The output of the candidate token in the self-attention module can be calculated as:
\begin{equation}
\small
    Q_c \leftarrow H_c W^c_Q \quad
K_c \leftarrow H_c W^c_K \quad
V_c \leftarrow H_c W^c_V \quad
\end{equation}
\begin{equation}
\small
    A_c \leftarrow \text{softmax}\left(Q_c \left( K \oplus K_c \right)^T  \right) 
\end{equation}
\begin{equation}
\small
O_c \leftarrow V_c {W^c_O}^T \quad
    V_c \leftarrow  A_c \left( V \oplus V_c \right)^T
\end{equation}

where $H_c \in \mathbb{R}^{ d_{dec}}$ is the input hidden state of the candidate token, $\oplus$ represents the concatenation of matrices, and $K,V$ correspond to the representations of the normal tokens in one chunk. 

\subsection{\textbf{Parallel Decoding}} \label{Parallel_Decoding}
Through the dynamic condensing process described above, we obtain one candidate token for each chunk. Notably, the process of obtaining the candidate token from each chunk is independent, enabling \textit{parallel forwarding} for all chunks. Then the key/value representations of the candidate tokens are concatenated with the tokens in the local context layer by layer, as shown in Figure \ref{fig:framework}, and are finally processed by a frozen decoder to generate the next token.

We formally define the process of simultaneously generating candidate tokens from different chunks and then aggregating these candidate tokens to produce the final token as \textit{parallel decoding}. This mechanism not only enables precise understanding of long contexts but also reduces the Transformer's original $O(L^2)$ computational complexity to $O((L/n)^2)$. A detailed efficiency analysis is provided in Appendix \ref{appendix_efficiency}.

\section{Training}\label{training}
Regarding training data, to ensure the generalizability of our method and maintain fairness in comparison with the baselines, we leverage RedPajama \cite{Together} as the training corpus and sample examples with sequence lengths varying between 3K and 8K tokens from it. RedPajama is an open-source pre-training dataset for LLaMA-1 \cite{touvron2023llama}, which is widely utilized in previous work \cite{zhang2024soaring,yen2024long}.  Detailed statistics are reported in Appendix \ref{appendix_training_data}.

\textbf{Auto-Regressive Loss.} Specifically, we train the model to predict the next token, and the loss is only applied to tokens in the local context, which encourages the candidate token to aggregate useful information from each chunk.
\begin{equation}
\footnotesize
\min_{F'_{\text{dec}}}  -\sum_{i=2}^{S-m} \log(p(x_{m+i} \mid c_1, \ldots, c_k, x_{m+1}, \ldots, x_{m+i-1})) 
\end{equation}
Here, $c_i$ represents the candidate token generated by the $i$-th chunk. Specifically, based on the relationship between the \textit{memory tokens} $\{x_1,...,x_m\}$ and the \textit{local context} $\{x_{m+1},...,x_S\}$, we design two loss functions for joint training. \textbf{i)} If the local context is a continuation of the memory tokens, we term this loss the \textit{Continuation Loss}, as it trains the model to naturally generate new tokens that follow the given context. \textbf{ii)} Alternatively, if we randomly select $L$ consecutive memory tokens as local context, we define this loss as the \textit{Reconstruction Loss}, as it trains the model to reconstruct tokens when clear contextual information is available. Subsequent experiments demonstrate that both types of loss are essential.
\begin{table*}[tb]
\centering
\small
\begin{tabular}{l|cccc|cccc|cccc}
\toprule
& \multicolumn{4}{c|}{PG19} & \multicolumn{4}{c}{Proof-Pile} & \multicolumn{4}{c}{CodeParrot} 
\\
\cmidrule(lr){2-5} \cmidrule(lr){6-9} \cmidrule(lr){10-13}
\textbf{Method} & 4K & 16K & 32K & 100K & 4K & 16K & 32K & 100K & 4K & 16K & 32K & 100K\\
\midrule
Llama-2-7B & 9.21 & $\tiny{>}10^3$ & $\tiny{>}10^3$ & OOM &  3.47 & $\tiny{>}10^3$ & $\tiny{>}10^3$ & OOM & 2.55 & $\tiny{>}10^3$ & $\tiny{>}10^3$ & OOM\\
PI & 9.21 & 19.5 & $\tiny{>}10^2$ & OOM & 3.47 & 5.94 & 33.7 & OOM & 2.55 & 4.57 & 29.33 & OOM\\
NTK & 9.21 & 11.5 & 37.8 & OOM & 3.47 & 3.65 & 7.67 & OOM & 2.55 & 2.86 & 7.68 & OOM \\
StreamingLLM & 9.21 & 9.25 & 9.24 & 9.32 & 3.47 & 3.51 & 3.50 & 3.55 & 2.55 & 2.60 & 2.54 & 2.56 \\ 
\midrule
AutoCompre.-6K & 11.8 & $\tiny{>}10^2$ & $\tiny{>}10^3$ & OOM & 4.55 & $\tiny{>}10^2$ & $\tiny{>}10^3$ & OOM & 5.43 & $\tiny{>}10^2$ & $\tiny{>}10^3$ & OOM \\
YaRN-128K & 6.68 & 6.44 & 6.38 & OOM & 2.70 & 2.47 & 2.41 & OOM & 2.17 & 2.04 & 2.00 & OOM \\
LongChat-32K & 9.47 & 8.85 & 8.81 & OOM & 3.07 & 2.70 & 2.65 & OOM & 2.36 & 2.16 & 2.13 & OOM \\
LongAlpaca-16K & 9.96 & 9.83 & $\tiny{>}10^2$ & OOM & 3.82 & 3.37 & $\tiny{>}10^3$ & OOM & 2.81 & 2.54 & $\tiny{>}10^3$ & OOM \\
LongLlama & 9.06 & 8.83 & OOM & OOM & 2.61 & 2.41 & OOM & OOM & 1.95 & 1.90 & OOM & OOM \\
Activation Beacon & 9.21 & 8.54 & 8.56 & 8.68 & 3.47 & 3.42 & 3.39 & 3.35 & 2.55 & 2.54 & 2.53 & 2.55 \\
\midrule
FocusLLM & 9.21 & 9.19 & 9.17 & 10.59 & 3.47 & 3.17 & 3.43 & 2.57 & 2.55 & 2.01 & 2.27 & 3.02 \\
\bottomrule
\end{tabular}
\caption{Language Modeling Assessment: perplexity analysis of various context scaling methods on the PG19, Proof-Pile, and CodeParrot. FocusLLM successfully maintains low perplexity on extremely long sequences.}
\label{tab:perplexity}
\end{table*}

\section{Experiments}
In this section, we will conduct a comprehensive evaluation of the effectiveness of FocusLLM, spanning both language modeling and a variety of downstream tasks. We refer readers to Appendix \ref{appendix_exp_details} for detailed experimental settings including hyperparameters due to space constraints.

\subsection{Long-context Language Modeling}\label{language modeling}
In this section, we evaluate FocusLLM on long-context language modeling benchmarks, with text lengths ranging from 4K to 128K tokens. \newline
\textbf{Datasets.} We perform the evaluation on three datasets: PG19 \cite{rae2019compressive}, Proof-Pile \cite{proofpile}, and CodeParrot \cite{codeparrot}. These three datasets encompass 100 long test cases related to books, arXiv papers, and code repositories, respectively. The results of baseline models are token from \cite{zhang2024soaring} for comparison. Following the setting of \cite{yen2024long}, as FocusLLM relies on the last decoder to perform generation, we calculate the perplexity on the last 256 tokens of each sequence, and for the 128K length, we filter out documents exceeding 128K tokens and evaluate 10 samples due to data scarcity and computational cost. \newline
\textbf{Model.} FocusLLM is based on LLaMA-2-7B (chat), hence the models for comparison are all on the same scale, 7B. The baseline models can be categorized into the following types: i) Methods focusing on the modification of \textbf{positional encoding}, including Positional Interpolation \cite{chen2023extending}, the NTK-Aware Scale ROPE\footnote{https://www.reddit.com/r/LocalLLaMA/comments/14lz7j5/\\ntkaware\_scaled\_rope\_allows\_llama\_models\_to\_have/}, and the training-free method StreamingLLM \cite{xiao2023efficient}, which is based on attention sinks. 
ii) \textbf{Fine-tuned methods} trained on long inputs, such as LongAlpaca-16K \cite{chen2023longlora}, LongChat-32K \cite{li2023long}, and YaRN-128K \cite{peng2023yarn}.
iii)\textbf{ Methods with designed structures} specifically for long contexts, including AutoCompressor-6K \cite{chevalier2023adapting}, LongLlama \cite{tworkowski2024focused} and Activation Beacon \cite{zhang2024soaring}. For instance, Activation Beacon achieves compression of long texts by training the model to represent the information of a regular text segment with a small number of beacon tokens. \newline
\textbf{Analysis.} The results are presented in Table \ref{tab:perplexity}. Here are several observations we can make:
(1) Compared to the basic LLaMA-2-7B model and some fine-tuning free methods, our model demonstrates superior performance. When extending the context length from 4K to longer, the perplexity becomes lower, indicating that information from a longer context can be effectively utilized.
(2) FocusLLM achieves comparable performance to fine-tuned full-attention methods. This result is notable because our model operates with significantly higher training efficiency. For instance, LongLlama is fine-tuned using 7B tokens with all parameters being trainable. In contrast, FocusLLM uses $1/10$ of the training budget and $1/3$ of the parameters.
(3) FocusLLM can maintain language modeling capabilities at lengths much longer than other models while retaining precise comprehension of the entire text. Although models like StreamingLLM and Activation Beacon can still achieve lower perplexity by compressing tokens, they are unable to recover the previous context information, which severely affects their capabilities in downstream tasks. In summary, FocusLLM achieves comparable language modeling performance with a small training cost.

\begin{table*}[t]
    \resizebox{\textwidth}{!}{
    \centering
    \small
    \begin{tabular}{cl|rrrrrr|rrr}
    \toprule
    & & \multicolumn{9}{c}{Vicuna-7B-v1.5 (4K)}  \\
    &     & Original & LChat & Vic-16K & Yarn-128K	& PI & NTK  & Stream &InfLLM & FocusLLM\\ \midrule
    
  \multirow{7}{*}{\begin{tabular}[c]{@{}l@{}}$\infty$-Bench \end{tabular}} 
    & Math.Find     
              & 11.71 & 9.43 & 13.43 &17.14& OOM & OOM & 6.00 & 11.14 & \textbf{11.71}\\
    & En.MC         
            & 30.13 & 24.45 & 34.06& 27.95 & OOM & OOM &  \textbf{32.31} & 31.44 &\textbf{32.31} \\
    & Code.Debug    
                    & 38.83 & 27.66 & 35.03 &22.59& OOM & OOM &  \textbf{46.19} & 34.26 & 28.43 \\ 
    & Retrieve.KV   
                    & 1.40 & 1.40 & 1.00 & 0.00&OOM & OOM & 0.00 &  0.60 &\textbf{12.40} \\
    & Retrieve.Number  
                       & 4.41 & 23.90 & 10.34 &56.61& OOM & OOM & 4.41 & 81.69 &\textbf{83.56} \\
    & Retrieve.PassKey 
                       & 5.08 & 28.64 & 15.25 &92.71& OOM & OOM &  4.92 & \textbf{99.15} & 95.76 \\ 
   \cmidrule{2-11}
    & Average       & 15.26 & 19.25 & 18.19 &36.17& -- & -- & 15.64 & 43.05 & \textbf{44.03} \\
    \midrule
  \multirow{17}{*}{\begin{tabular}[c]{@{}l@{}}LongBench\end{tabular}} 
    & NarrativeQA   & 11.19 & 20.35 & 17.85 & 19.67& 0.78 & 5.66  &  15.61 & 15.53&\textbf{21.14}\\
    & Qasper        & 13.79 & 29.35 & 25.85 &11.10& 2.71 & 21.17 &  23.84 & 23.57 &\textbf{31.07}	\\
    & MultiFieldQA
    & 22.08 & 42.55 & 37.15 &35.06& 1.01 & 36.76 & 32.80 & \textbf{37.14}	&36.73	 \\
    & HotpotQA      & 12.71 & 33.19 & 24.72 &11.94& 1.35 & 19.54 &  22.17 & 22.53 & \textbf{40.65}\\
    & 2WikiMQA      & 13.99 & 24.33 & 21.41 &12.02& 1.17 & 14.51 & 18.38 & 18.82 &\textbf{20.30}	\\
    & Musique       & 4.81  & 14.71 & 8.44  &7.52& 0.71 & 4.30   & 6.30 & 5.24	& \textbf{14.20}	\\
    & GovReport     & 27.67 & 30.83 & 27.62 &29.46& 1.9  & 25.26 &  23.18 & \textbf{26.79}	&26.66 \\
    & QMSum         & 19.72 & 22.93 & 22.63 &21.53& 1.29 & 19.48 &  20.09 & \textbf{20.91}	& 20.50\\
    & MultiNews     & 26.61 & 26.63 & 27.88 &16.04& 1.16 & 25.88 & 26.19 & 26.43	&\textbf{27.45}\\
    & TREC          & 69.00 & 66.50 & 69.00 & 68.50& 4.50 & 59.00 & 61.00 & \textbf{67.50}	& \textbf{68.00} \\
    & TriviaQA      & 81.94 & 83.99 & 85.63 &88.21& 0.90 & 25.85 &  78.81 & \textbf{84.36} &81.63 \\
    & SAMSum        & 35.12 & 12.83 & 9.15  &26.52& 0.12 & 5.05  &  32.46 & 31.89	&\textbf{35.36}	\\
    & PassageRetrieval
        & 9.00  & 30.50 & 4.00 &16.25 & 0.62  & 5.00  & 6.00 & 9.00	& \textbf{15.67} \\
    & LCC           & 64.53 & 54.79	& 50.64 &66.39& 21.54 & 53.65 & \textbf{ 63.70} & 61.41 &62.79	\\
    & RepoBench-P   & 50.17 & 58.99 & 44.94 &55.82& 19.36 & 44.58 & 48.26 & 47.52 & \textbf{53.72}	\\ 
    \cmidrule{2-11}
    & Average       & 30.82 & 34.70 & 31.79 &32.40& 3.94 & 24.38 & 31.92 & 33.24&\textbf{36.17} \\ 
    \bottomrule
    \end{tabular}
    }
    \caption{The results on $\infty$-Bench and LongBench.  The models on the right part can process extremely long inputs. On both benchmarks, FocusLLM achieves significant improvements compared to strong baselines. }
    \label{tab:vicuna-results}
\end{table*}

\subsection{Downstream Tasks}\label{downstream}
\textbf{Datasets.}
To assess the capabilities of FocusLLM in real-world scenarios, we select two widely used datasets: Longbench \cite{bai2023longbench} and $\infty$-Bench \cite{zhang2024inftybench}.
Longbench offers an evaluation on a variety of tasks including question answering, summarization, few-shot learning, mathematical counting, and code completion. $\infty$-Bench is designed to test a model's ability to understand and reason over super long contexts, with an average length of 145.1K tokens. Thus, the tasks in $\infty$-Bench are well-suited to test whether the model has a precise understanding of long contexts without \textbf{\textit{information loss}}.
For more detailed statistics, please refer to Appendix \ref{Details of Benchmarks}. We believe that these two benchmarks can comprehensively reflect the capabilities of the model on downstream tasks. \newline
\textbf{Models.}
We select representative models from the three types of baselines mentioned in Section \ref{language modeling} for comparison. Additionally, we focus on comparing FocusLLM with recently proposed models capable of processing extremely long streaming inputs. Specifically, StreamingLLM utilizes a sliding window mechanism; InfLLM \cite{xiao2024infllm} stores processed context into memory units and retrieves it using attention scores; Activation Beacon compresses the preceding text to maintain a smaller context length. CEPE \cite{yen2024long} 
 adopts a small encoder to process long inputs chunk by chunk and feeds the memory to a decoder by cross-attention. \newline
\textbf{Main Results.} The experimental results are displayed in Table \ref{tab:vicuna-results} and \ref{tab:llama-results}. We reference some baseline results from \cite{xiao2024infllm}, which are based on the Vicuna-7B-v1.5 model. Vicuna-7B-v1.5 is based on LLaMA-2-7B but fine-tuned on conversational data. For a fair comparison, we also train a Vicuna version of FocusLLM. For YaRN-128K, we select the version based on Mistral-7B-inst-v0.2, which is stronger than Vicuna. For LongLlama, as they do not have a version based on the Llama2, we directly utilize the officially released model. CEPE and LongLLaMA will experience \textit{OOM} on $\infty$-Bench due to their substantial memory usage, so we only report their results on LongBench. Since not all models are inherently capable of processing infinite text lengths, we also elaborate the effective lengths for each method presented in Tables \ref{tab:vicuna-results} and \ref{tab:llama-results} in Appendix \ref{effective length}.

From the experimental results, we can make the following comparisons between FocusLLM and previous methods: \textbf{(1)} FocusLLM outperforms all baseline models, achieving \textbf{\textit{the best results}} on both the relatively shorter benchmark Longbench and the extremely long benchmark $\infty$-Bench. This demonstrates FocusLLM's capability for effective understanding and reasoning on long sequences and its broad applicability. \textbf{(2)} Different types of baseline models exhibit various shortcomings. For training-free models like PI and NTK, extending the length to 128K comes with a significant sacrifice in performance. Due to the lack of precise understanding of the full context, models that employ sliding window or condensing techniques, such as StreamingLLM and Activation Beacon perform poorly on $\infty$-Bench (see also Appendix \ref{appendix_ab_on_infinite_bench}), with performance nearly approaching zero on some tasks. This indicates that \textbf{\textit{they suffer from severe information loss}}. As for fine-tuned models like LongChat and CEPE, their limitation is the restricted supported length. For example, CEPE struggles to handle lengths beyond 128K effectively \cite{yen2024long}. \textbf{(3)} The approaches of length extrapolation and continual training on long inputs, while capable of scaling context, introduce substantial computational and memory costs. In contrast, FocusLLM processes the text in chunks and utilizes parallel decoding, which significantly conserves both the memory and time for inference.

\begin{table}[t]
\resizebox{.5\textwidth}{!}{
    \centering
    \footnotesize
    \begin{tabular}{l|rrrrr}
    \toprule 
     & \multicolumn{5}{c}{Llama-7B-chat (4K)}  \\
         & Original & CEPE & L\_L & A\_B & FocusLLM  \\ \midrule
  
     NarrativeQA   & 18.70  & 22.14 & - & -  & 20.38 \\

    Qasper        & 19.20  & 26.34  & -& -  & 21.73 \\

    MultiFieldQA
                 & 36.80  & 31.56  & -& -  & 36.91 \\
                    
    \quad -Average 
                    & 24.90 & 26.68 & \textbf{30.12}&27.14 & 26.34 \\
    \cmidrule{1-6}

 HotpotQA      & 25.40  & 34.95 & - & -  & 38.95 \\

    2WikiMQA      & 32.80  & 32.39 & - & -  & 32.95 \\

 Musique       & 9.40   & 9.76  & - & -  & 15.39 \\
    
     \quad -Average  
          & 22.60 & 25.70 &16.37 & 28.28 & \textbf{29.10} \\
    \cmidrule{1-6}

 GovReport     & 27.30  & 13.90  & - & -  & 25.54 \\

    QMSum         & 20.80  & 20.30  & - & -  & 21.86 \\ 

    MultiNews     & 25.80  & 3.10  & - & -  & 26.35 \\

     \quad -Average 
                    & 24.70 & 12.43 & 24.19&\textbf{25.15} & 24.55 \\
    \cmidrule{1-6}

 TREC          & 61.50  & 68.50  & - & -  & 68.00  \\

 TriviaQA      & 77.80  & 87.90  & - & -  & 85.08 \\
    SAMSum        & 40.70  & 32.38  & -& -  & 41.63 \\

    \quad -Average 
                    & 60.00  & 62.92 &60.31& 60.72 & \textbf{64.81} \\
    \cmidrule{1-6}
	
     LCC           & 52.40  & 66.21  & -& -  & 58.42 \\

 RepoBench-P   & 43.80  & 58.94  & -& -  & 54.27 \\

     \quad -Average 
                    & 48.10 & 62.57 & \textbf{66.05}&57.83 & 56.35  \\

    \cmidrule{1-6}
    Average       & 35.20 & 36.31 & 37.50 & 38.54 & \textbf{39.01} \\
    
    \bottomrule
    \end{tabular}
    }
    \caption{The results of LLaMA2-based models on tasks of LongBench. L\_L represents Long Llama and A\_B represents Activation Beacon. FocusLLM outperforms memory-based and compression-based methods, and maintains attention to all tokens of context.}
    \label{tab:llama-results}
\end{table}

\section{Further Exploration}
\subsection{Visualization of Candidate Tokens}
To further illustrate how candidate tokens function, we provide a more intuitive explanation by visualizing the information carried by these tokens through attention weight heatmaps when decoding the next token. Due to space limitations, we place the visualization results in Appendix \ref{appendix_visualize}. We have the following observations: i) In Passkey Retrieval task, the model assigns a high attention weight to \textit{one certain candidate token}, indicating that this token effectively carry the passkey information from its respective chunk. In contrast, candidate tokens from chunks containing noisy text carry no useful information, resulting in near-zero attention weights. ii) In LongBench NarrativeQA task, the model shows a slightly different pattern, where \textit{many candidate tokens receive attention}, as multiple chunks' information may be aggregated for the QA task. The visualization results demonstrate that FocusLLM effectively uses candidate tokens to transmit information from the context while ignoring irrelevant noise.
\subsection{Scaling to 400K Context}
We contend that FocusLLM is capable of processing extremely long sequences. To validate this, we first conduct experiments on the passkey retrieval task \cite{mohtashami2024random}. The results, as illustrated in Figure \ref{fig:passkey}, demonstrate that FocusLLM maintains nearly 100\% effectiveness at lengths of up to 400K\footnote{Constrained by hardware, the maximum length we are able to test is 400k tokens.}, outperforming all other models. We also extended the language modeling experiments introduced in Section \ref{language modeling} to 400K, a length at which most models fail to manage effectively. The result is presented in the Appendix \ref{appendix_400k}.


\begin{figure*}[ht]
    \centering
    \begin{minipage}[t]{0.32\textwidth}
    \vspace{-\baselineskip}
        \includegraphics[width=\textwidth]{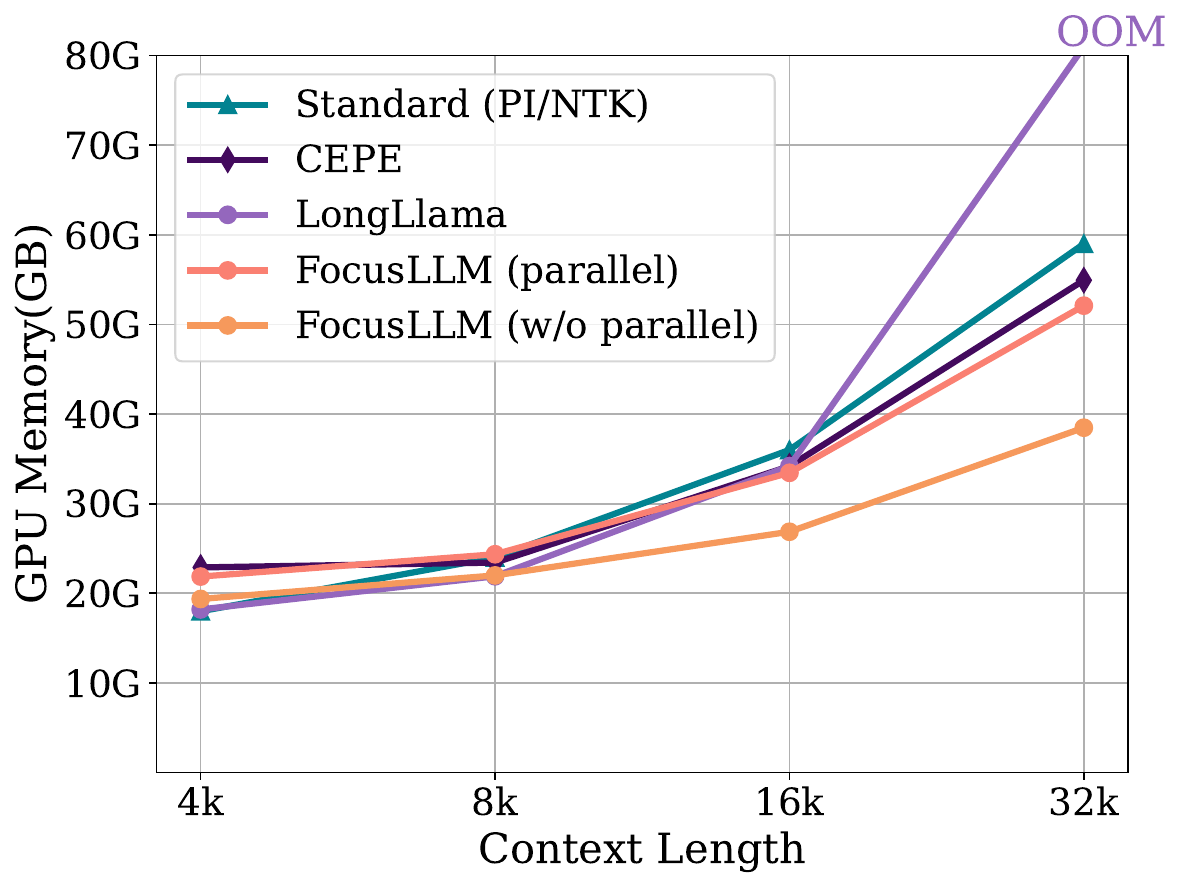} 
        \caption{FocusLLM exhibits a more efficient growth pattern in memory usage compared to previous methods.}
        \label{fig:memory_usage}
    \end{minipage}\hfill
    \begin{minipage}[t]{0.32\textwidth}
    \vspace{-\baselineskip}
        \includegraphics[width=\textwidth]{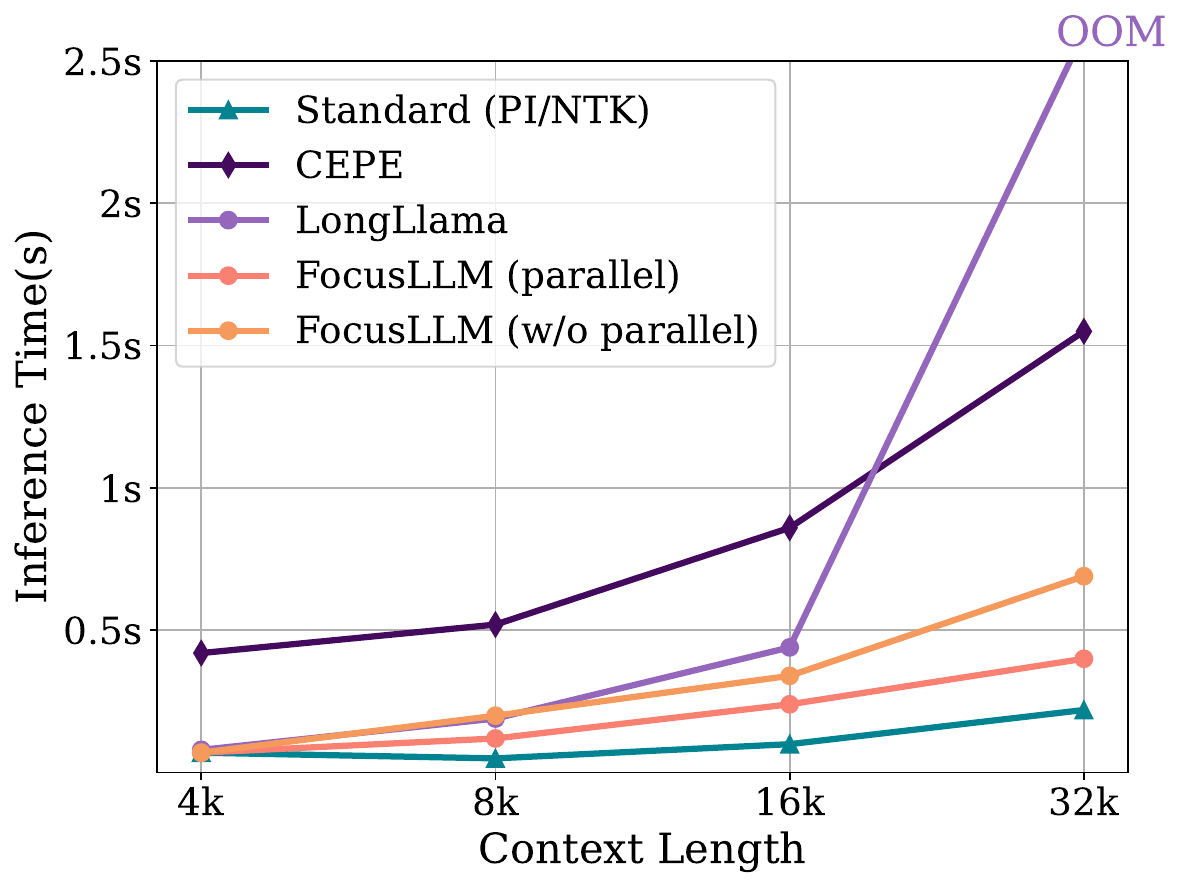} 
        \caption{Comparison of inference time. The time taken by FocusLLM is superior to previous methods.}
        \label{fig:inference_time}
    \end{minipage}\hfill
    \begin{minipage}[t]{0.32\textwidth}
    \vspace{-\baselineskip}
    \includegraphics[width=\textwidth]{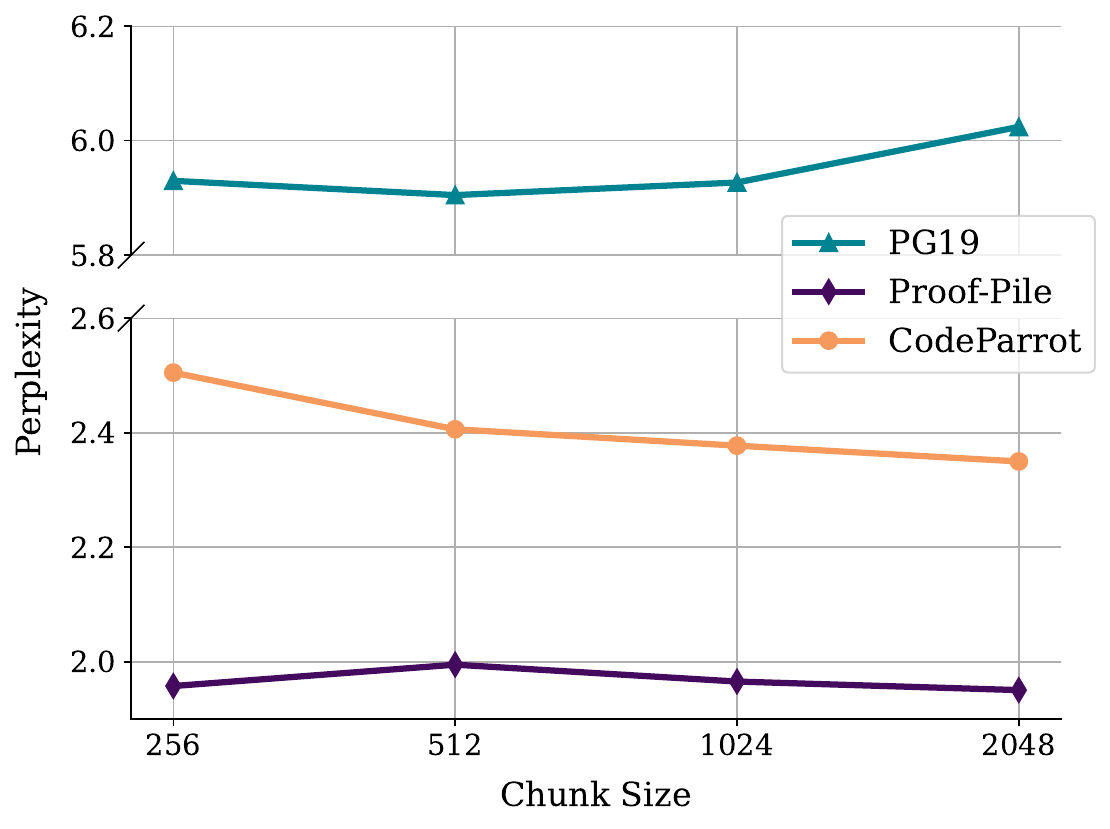}
    \caption{Perplexity under different chunk size with the total sequence length fixed as 8K on three datasets.}\hfill
    \label{fig:chunk size}
    \end{minipage}
\end{figure*}

\subsection{Memory Footprint and Inference Time}
For models that focus on long texts, aside from training costs, another critical aspect is the memory footprint and inference time. In this section, we compare FocusLLM with several previous long-context methods capable of \textit{retaining global information by preserving the cache} of all context: Standard (PI/NTK), LongLlama, and CEPE. As for models like Activation Beacon and StreamingLLM, although they maintain a constant memory footprint by only retaining cache for a fixed window, they suffer significant \textit{information loss} and struggle with the precise understanding of long texts as demonstrated in Section \ref{downstream}. Therefore, they are not the primary subjects of comparison.

The results are shown in Figure \ref{fig:memory_usage} and Figure \ref{fig:inference_time}.  \textit{FocusLLM with or without parallel} indicates whether we process each chunk either concurrently or sequentially. The findings indicate that: (1)  When ample memory resources are available, parallel processing is more efficient for FocusLLM. (2) Although FocusLLM splits long texts into numerous chunks, resulting in a slightly longer inference time compared to the standard approach, it still holds a significant advantage over other long-context methods.
\subsection{Chunk Size}
We conduct an investigation into the impact of different chunk sizes on performance. In theory, larger chunk sizes, as long as they do not exceed the model's default context length (e.g., 4K for LLaMA-2), are preferable because they allow for processing the memory with a smaller number of forward passes. However, smaller chunk sizes may enable more precise processing. 

In experiments, we maintain a total sequence length of 8K, testing the perplexity using different chunk sizes on the same samples of PG19. We select \{256, 512, 1024, 2048\} as our test sizes. The results are shown in Figure \ref{fig:chunk size}. We observe that there is no consistent trend in perplexity as the chunk size increases; it remains relatively stable. This confirms our hypothesis that we can employ larger chunk sizes on models with longer default context lengths (e.g. LLaMA-2-32K). We will explore this direction in our future work.

\begin{table*}[htbp]
    \centering
    \small
        \begin{tabular}{llccccc}
            \toprule
             & & \multicolumn{2}{c}{\textbf{LongBench}} & \multicolumn{3}{c}{\textbf{$\infty$-Bench}} \\ 
            \cmidrule(lr){3-4} \cmidrule(lr){5-7}
            & Hyper Params.  & NarrativeQA & TREC & Math.Find & En.MC & Retrieve.PassKey \\
            \midrule
    FocusLLM    & (2K, 2K) & \textbf{18.53} & \textbf{65.5}  & 13.43 & \textbf{31.00} & \textbf{99.32} \\
    \midrule
    Continuation Loss only & (2K, 2K) & 17.36 & 60.5 & \textbf{13.71}   & 27.95 & 1.69  \\
    Reconstruction Loss only & (2K, 2K) & 17.05 & 62.0  & 12.86 & 26.64 & \textbf{91.19}  \\
    \midrule
    Local Context Size $ \downarrow$ &  (1K, 2K) & 17.87 & 63.0 &8.86 & 29.69 & \textbf{99.32} \\
    \bottomrule
        \end{tabular}
\caption{
     Investigations into the training loss and local context size of FocusLLM. We present the results for representative tasks from LongBench and $\infty$-Bench. For instance, NarrativeQA belongs to Single-Doc QA, while TREC relates to Few-shot learning. The Hyper Params is denoted as (local context size, chunk size).
    }
    \label{tab:ablation}
\end{table*}

\subsection{Ablation Studies}
We employ both Continuation Loss and Reconstruction Loss for the training of FocusLLM. The motivation behind this is to equip the model with the natural language modeling capability while also enhancing its ability to recover information.
Ablation studies as detailed in Table \ref{tab:ablation}, reveal that relying solely on the Continuation Loss enables the model to manage some tasks effectively. Nonetheless, for tasks with substantial dependencies on the preceding context, like HotpotQA and Retrieve.PassKey, the model's efficacy deteriorates. Similarly, while employing the Reconstruction Loss ensures accurate restatement of the preceding context, the lack of generalizability of generating new tokens leads to a considerable decrease in performance. Therefore, the combined use of both loss functions is crucial for enhancing the performance and generalizability of FocusLLM.

We also investigate how the local context size influences performance in the last row of Table \ref{tab:ablation}. As we reduce the local context size from 3.5K to 1K, the performance of most tasks experiences a slight decline. This suggests that candidate tokens cannot fully replace the information within the context.

\section{Related Work}
\subsection{Long-context language models}
 One research direction involves length extrapolation in transformers \cite{peng2023yarn,jin2024llm}, where methods like positional interpolation help models adapt to longer sequences \cite{chen2023extending}. However, these techniques often fail to address the distraction issue caused by noisy content within extended texts \cite{tworkowski2024focused}. Another research branch focus on modifying the attention mechanism or employing compression techniques to maintain long texts within manageable lengths \cite{chevalier2023adapting,zhang2024soaring}. For instance, \cite{xiao2023efficient} discovered that retaining `sink tokens' in conjunction with a sliding window can achieve smooth streaming output. \cite{zhang2024soaring} expanded the context dramatically through compression. However, these methods share a common limitation: they cannot utilize information from all tokens. 
\subsection{Memory-enhanced Model}
The integration of memory layers within transformer architectures has become a pivotal strategy for enhancing long-context comprehension \cite{bertsch2024unlimiformer,tworkowski2024focused}. Common methodologies in memory-enhanced models often employ recurrent strategies that iteratively integrate information from the current window into a persistent memory \cite{munkhdalai2024leave}. Another approach is to initially encode the complete long text into memory tokens, which is then queried in to retrieve pertinent information as needed \cite{xiao2024infllm}. For example, \cite{yen2024long} employ a small encoder to sequentially encode long text segments, followed by the integration of these encoded chunks into a decoder. However, the drawback of such methods is that the memory length does not extrapolate well, and expanding the memory still incurs substantial computational costs. In contrast, FocusLLM offers superior training efficiency and remains effective on exceedingly long texts. 

\section{Conclusion}
In this work, we introduced FocusLLM, a novel framework that significantly extends the context length of LLMs. The core innovation lies in the parallel decoding strategy, which distribute the burden of understanding long texts across each chunk and effectively aggregating global information. FocusLLM stands out due to its  remarkable training efficiency, allowing us to achieve substantial gains in context comprehension with minimal computational and memory cost. Compared to existing methods, FocusLLM not only exhibits superior performance across downstream tasks but also maintains low perplexities when handling extensive texts, up to 400K tokens. We hope FocusLLM can be an inspiring work for the community, driving further exploration of long-context models.

\section{Limitations}
Our research has certain limitations:  (1) Due to hardware constraints, our tests were limited to 400K tokens, which does not represent the upper bound of FocusLLM's capabilities. Future work will explore the full performance potential of FocusLLM and investigate the use of quantization methods to reduce operational costs.
(2) While FocusLLM demonstrates exceptional training efficiency, we have observed that training on larger datasets can significantly enhance its generalizability and performance. Therefore, increasing the training data size will be a focus of future research.
\bibliography{custom}

\appendix

\section{\textbf{Efficiency of FocusLLM}} 
\label{appendix_efficiency}
The parallel decoding mechanism of FocusLLM effectively reduces the computational complexity of the standard architecture. Specifically, when dealing with very long sequences, the primary computational burden in the transformer architecture lies in the attention mechanism, which has a complexity of $O(L^2)$, where $L$ represents the total sequence length. By dividing the sequence into $n$ chunks, the complexity within each chunk becomes $O((L/n)^2)$. Therefore, when we process chunks in parallel, the time complexity can be reduced to \( O((L/n)^2) \). And the space complexity of $n$ chunks becomes 
 approximately $O((L/n)^2 * n) = O(L^2/n)$. This means that compared to a standard transformer, FocusLLM can reduce the computational complexity to a fraction, $1/n$ or even more of the original theoretically, where $n$ is the number of chunks into which the sequence is divided. In experiments, the longer the sequence length, the more apparent the improvement in efficiency.
\section{Details of Training Data}
\label{appendix_training_data}
We randomly sampled 80K sequences from RedPajama as our training corpus.
Table \ref{tab:length} shows the detailed distribution.
\begin{table}[htb]
    \centering
    \begin{tabular}{l|ccc|c}
        \toprule
        Length  & 3K$\sim$4K & 4K$\sim$6K & 6K$\sim$8K & Total\\
        \midrule
        Count  & 30K & 16K & 34K & 80K\\
        Portion & 38\% & 20\% & 42\% & 100\%\\
        \bottomrule
    \end{tabular}
        \caption{Length distribution of training corpus.}
    
    \label{tab:length}
\end{table}
\section{Experimental Details}
\label{appendix_exp_details}
We primarily conduct experiments on the LLaMA2-7B-Chat model. The additional trainable parameters mentioned in Section \ref{architecture} amount to only 2B approximately. 

Specifically, we conducted training on a Linux server equipped with 8×A100 GPUs, each with 40GB of memory. The training was carried out for 10,000 steps, equivalent to one epoch of the entire training dataset, using a batch size of 8 and a learning rate of 5e-5 with a linear scheduler. To conserve GPU memory, we employed deepspeed's zero2\_offload optimizing stage. The training process was completed in approximately 20 hours.

 For hyper-parameters, during training, the chunk size was randomly selected from the set \{64, 128, 256, 1024, 2048\}. For the length of tokens injected into each chunk, we set a default of 512 tokens for inference. And we ensured this length did not exceed the chunk size in the training procedure. As a result, the length of injected tokens was $\min\{512, chunk\ size\}$.  For evaluations on the Longbench, we adopt a larger local context size of 3,500 tokens for FocusLLM, consistent with the official setting. 
\section{Details of Benchmarks} \label{Details of Benchmarks}
\begin{table*}[!ht]
    \centering
    \small
\begin{tabular}{c|c|c|c|c|c}
    \hline
        Task & Task Type & Eval metric & Avg len & Language &  Sample \\ \hline
        HotpotQA & Multi-doc QA & F1 & 9,151 & EN & 200 \\ \hline
        2WikiMultihopQA & Multi-doc QA & F1 & 4,887 & EN & 200 \\ \hline
        MuSiQue & Multi-doc QA & F1 & 11,214 & EN & 200 \\ \hline
        MultiFieldQA-en & Single-doc QA & F1 & 4,559 & EN & 150 \\ \hline
        NarrativeQA & Single-doc QA & F1 & 18,409 & EN & 200 \\ \hline
        Qasper & Single-doc QA & F1 & 3,619 & EN & 200 \\ \hline
        GovReport & Summarization & Rouge-L & 8,734 & EN & 200 \\ \hline
        QMSum & Summarization & Rouge-L & 10,614 & EN & 200 \\ \hline
        MultiNews & Summarization & Rouge-L & 2,113 & EN & 200 \\ \hline
        TriviaQA & Few shot & F1 & 8,209 & EN & 200 \\ \hline
        SAMSum & Few shot & Rouge-L & 6,258 & EN & 200 \\ \hline
        TREC & Few shot & Accuracy & 5,177 & EN & 200 \\ \hline
        PassageRetrieval-en & Synthetic & Accuracy & 9,289 & EN & 200 \\ \hline
        LCC & Code & Edit Sim & 1,235 & Python/C\#/Java & 500 \\ \hline
        RepoBench-P & Code & Edit Sim & 4,206 & Python/Java & 500 \\ \hline
    \end{tabular}
    \caption{ Detailed statistics of the tasks used in our paper of LongBench. }
    \label{appendix_longbench}
\end{table*}

\begin{table*}[!ht]
    \centering
    \small
    \begin{tabular}{c|c|c|c|c}
    \hline
        Task Name & Context &  Examples & Avg Input Tokens & Avg Output Tokens \\ \hline
        En.MC & Fake Book & 229 & 184.4k & 5.3  \\ \hline
        Code.Debug & Code Document & 394 & 114.7k & 4.8 \\ \hline
        Code.Run & Synthetic & 400 & 75.2k & 1.3  \\ \hline
        Math.Find & Synthetic & 350 & 87.9k & 1.3 \\ \hline
        Retrieve.PassKey & Synthetic & 590 & 122.4k & 2.0  \\ \hline
        Retrieve.Number & Synthetic & 590 & 122.4k & 4.0  \\ \hline
        Retrieve.KV[\^2] & Synthetic & 500 & 89.9k & 22.7 \\ \hline
    \end{tabular}
    \caption{ Detailed statistics of the tasks used in our paper of $\infty$-Bench.}
    \label{appendix_infbench}
\end{table*}

\subsection{LongBench}
LongBench\cite{bai2023longbench} includes 14 English tasks, 5 Chinese tasks, and 2 code tasks, with the average length of most tasks ranging from 5K to 15K. In experiments, we only utilize the English tasks. Detailed statistics of the tasks used in our paper are shown in Table \ref{appendix_longbench}.

\subsection{$\infty$-Bench}

The benchmark \cite{zhang2024inftybench} comprises 12 unique tasks, each crafted to assess different aspects of language processing and comprehension in extended contexts. Detailed statistics of the tasks used in our paper are shown in Table \ref{appendix_infbench}.

\section{Details of the effective lengths of models in Table \ref{tab:vicuna-results} and \ref{tab:llama-results}}\label{effective length}
Not all models are capable of processing infinite text lengths. Therefore, we provide a clear explanation of the effective input length for each method in Table \ref{tab:vicuna-results} and Table \ref{tab:llama-results}.
Specifically: (i) For models with a finite context length, we truncate the inputs by only preserving the system prompts and the tail of inputs to simulate real-world applications with streaming inputs like \cite{xiao2024infllm}. For instance, in Table \ref{tab:vicuna-results}, these models include Original (4K), LChat (32K), Vic-16K (16K), Yarn (128K), PI (128K), and NTK (128K). (ii) For other models, including StreamingLLM, InfLLM, LongLlama, CEPE, Activation Beacon, and our FocusLLM, the input can theoretically be of any length. So  we input the entire sequence on the two benchmarks.
\begin{table}[t]
    \centering
    \small
    \begin{tabular}{c|c}
    \hline
        ~  & Activation Beacon  \\ \hline
        Code Debug  & 21.32  \\ \hline
        Math Find  & 11.71  \\ \hline
        Math Calc  & 0.00  \\ \hline
        Passkey  & 1.69 \\ \hline
        Number String & 1.69\\ \hline
        KV Retrieval  & 0.00 \\ \hline
    \end{tabular}
    \caption{The accuracy of Activation Beacon on $\infty$-Bench.}
    \label{ab_on_inf}
\end{table}

\begin{figure}[t]
\centering
   \includegraphics[width=0.3\textwidth]{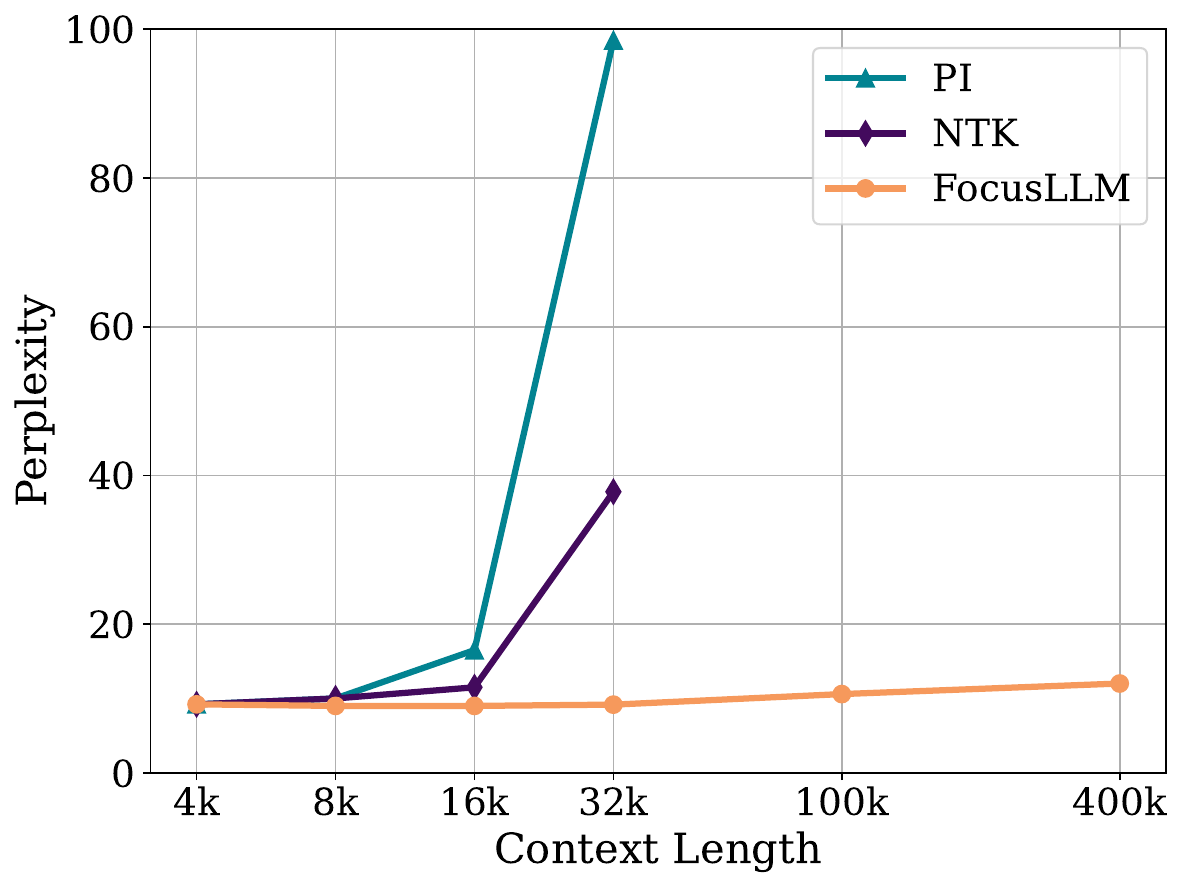} 
        \caption{Perplexity on PG19 dataset of FocusLLM compared to methods PI and NTK. FocusLLM can maintain low perplexity even at token counts up to 400K tokens.}
        \label{fig:perplexity}
\end{figure}
\section{Supplementary Results on $\infty$-Bench of Activation Beacon}\label{appendix_ab_on_infinite_bench}
Due to the compression of the context cache, Activation Beacon cannot retain full global information, which hinders its ability to handle tasks that require precise comprehension of the entire text in real-life scenarios, as demonstrated in the results presented in the Table \ref{ab_on_inf}.

\section{Scaling language modeling to 400K context}\label{appendix_400k}
As shown in Figure \ref{fig:perplexity}, FocusLLM maintains a low perplexity even with a context length of 400K. Note that the number of candidate tokens corresponding to 400K is 200, which is far greater than the number of candidate tokens seen during training. This demonstrates that FocusLLM has strong extrapolation capabilities. We can effectively scale to lengths greater than 400K by either using longer sequences during training or by employing a base model with a default context length, which we plan to explore in future work.

\section{Visualization of Attention Heatmap}\label{appendix_visualize}
\begin{figure*}[ht]
    \centering
    \begin{minipage}[b]{0.28\textwidth} %
        \centering
        \includegraphics[width=\textwidth]{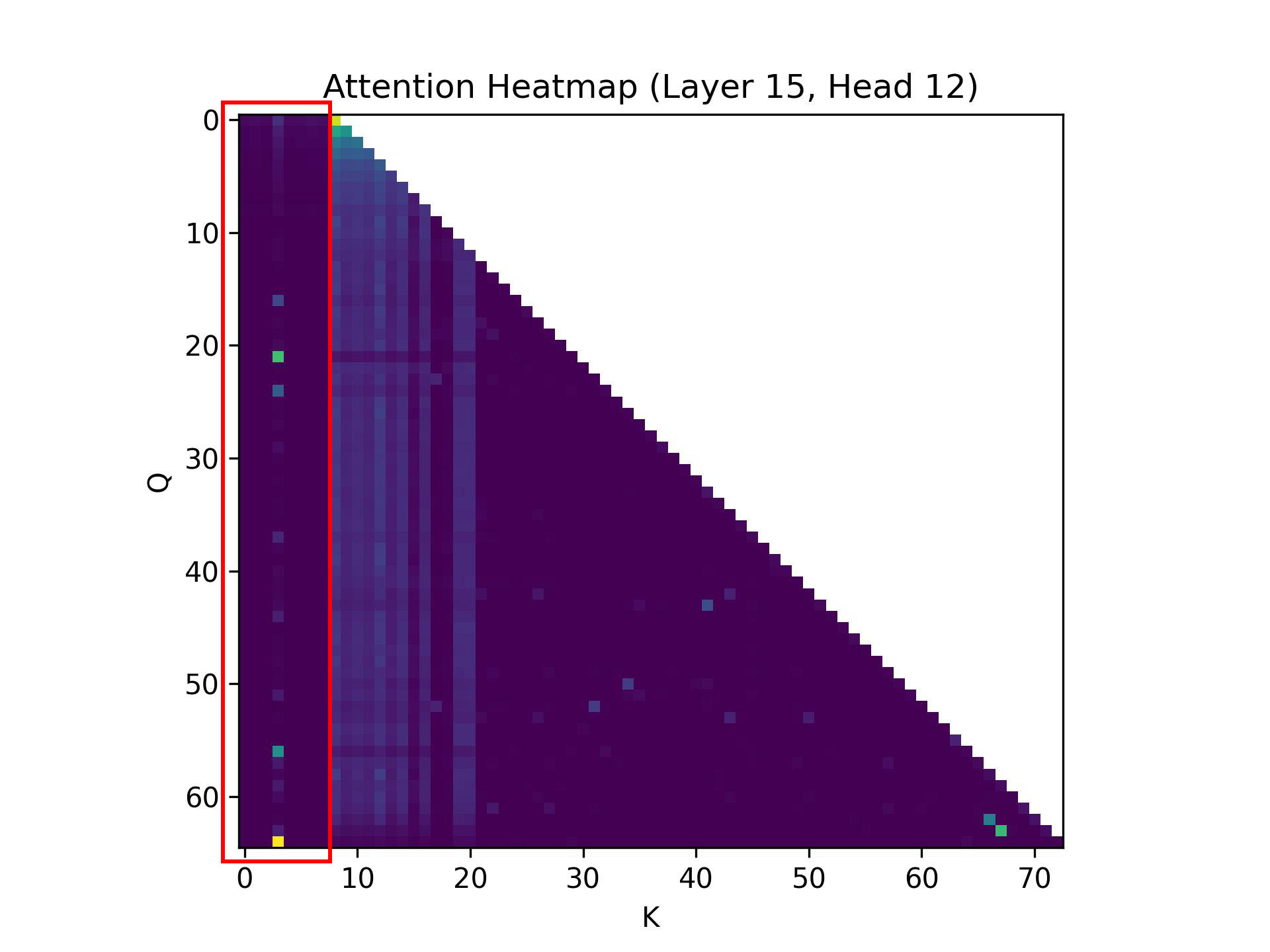} %
        \label{fig:heatmap}
    \end{minipage}\hfill
    \begin{minipage}[b]{0.28\textwidth} %
        \centering
        \includegraphics[width=\textwidth]{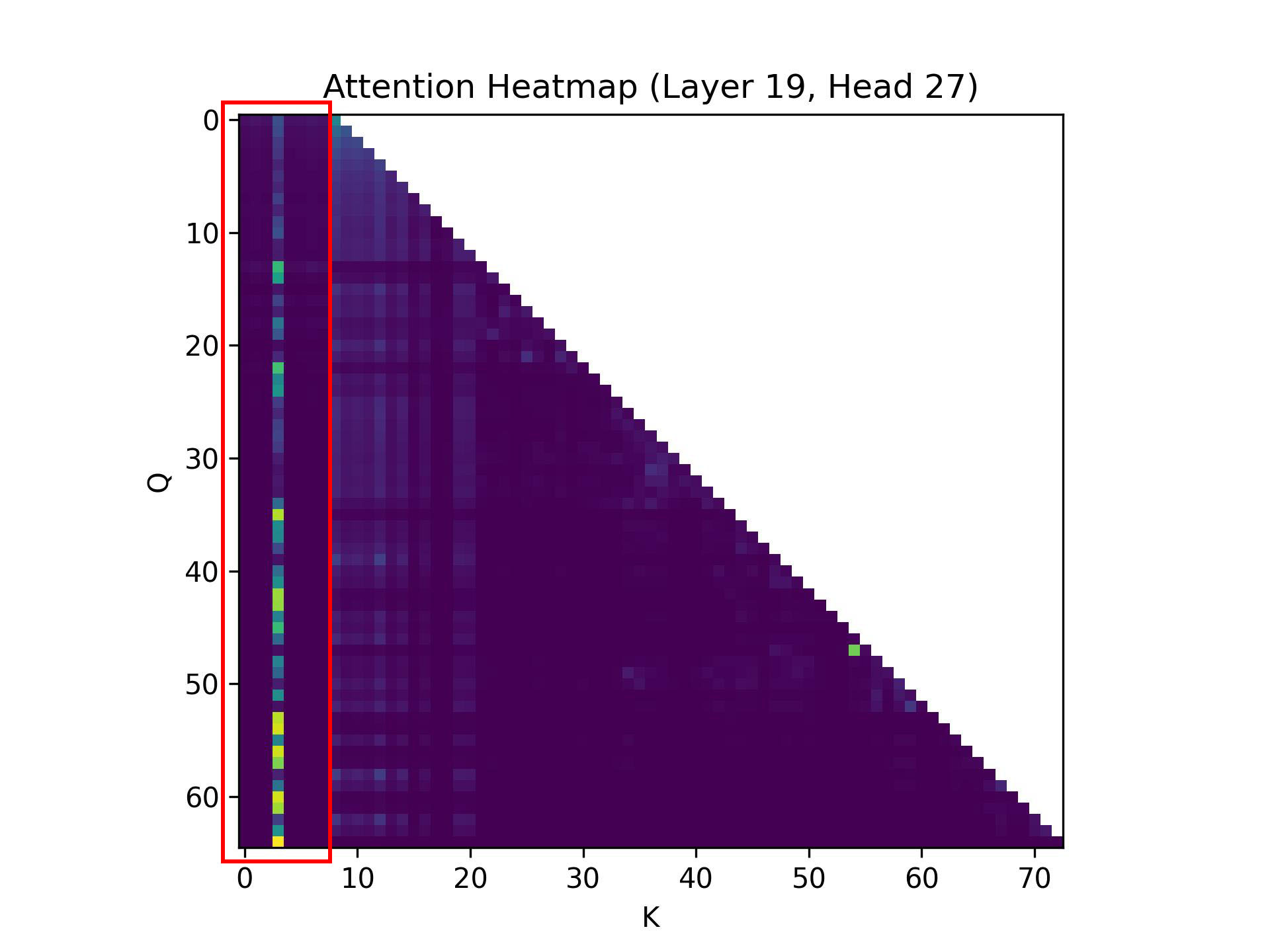} %
        \label{fig:heatmap}
    \end{minipage}\hfill
    \begin{minipage}[b]{0.28\textwidth}
        \centering
        \includegraphics[width=\textwidth]{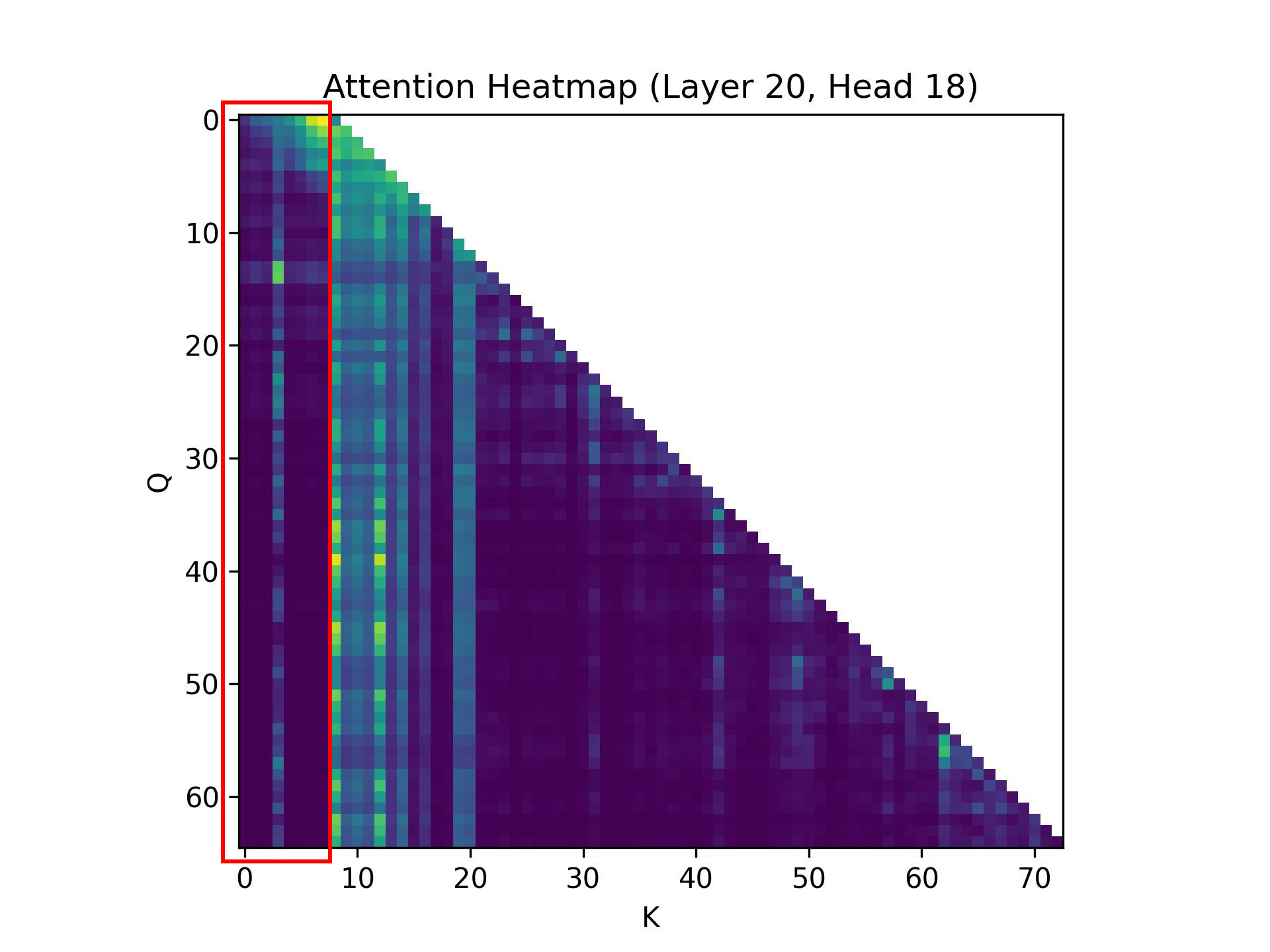} %
        \label{fig:image2}
    \end{minipage}\hfill
    \\
    \begin{minipage}[b]{0.28\textwidth} %
        \centering
        \includegraphics[width=\textwidth]{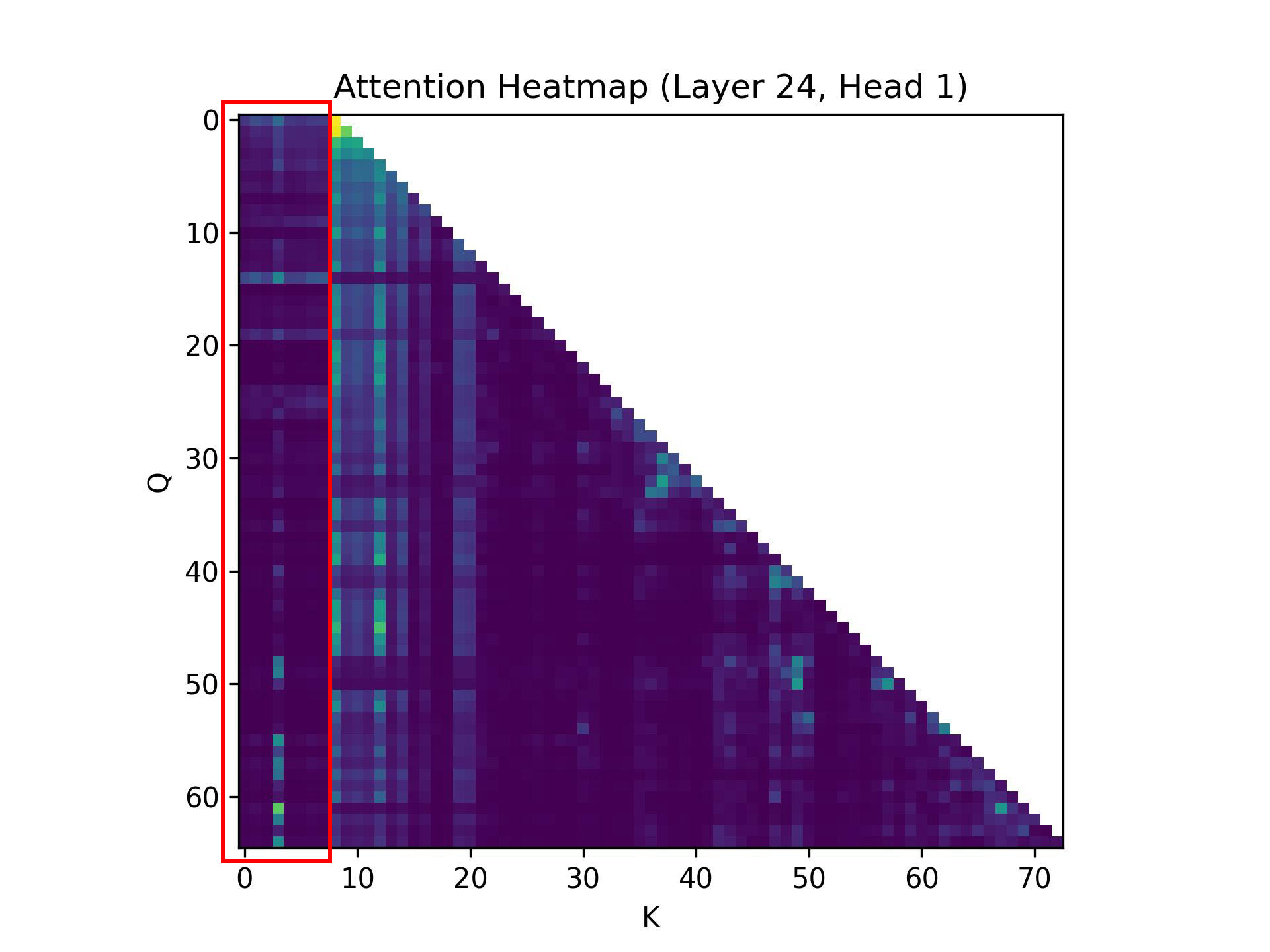} %
        \label{fig:heatmap}
    \end{minipage}\hfill
    \begin{minipage}[b]{0.28\textwidth}
        \centering
        \includegraphics[width=\textwidth]{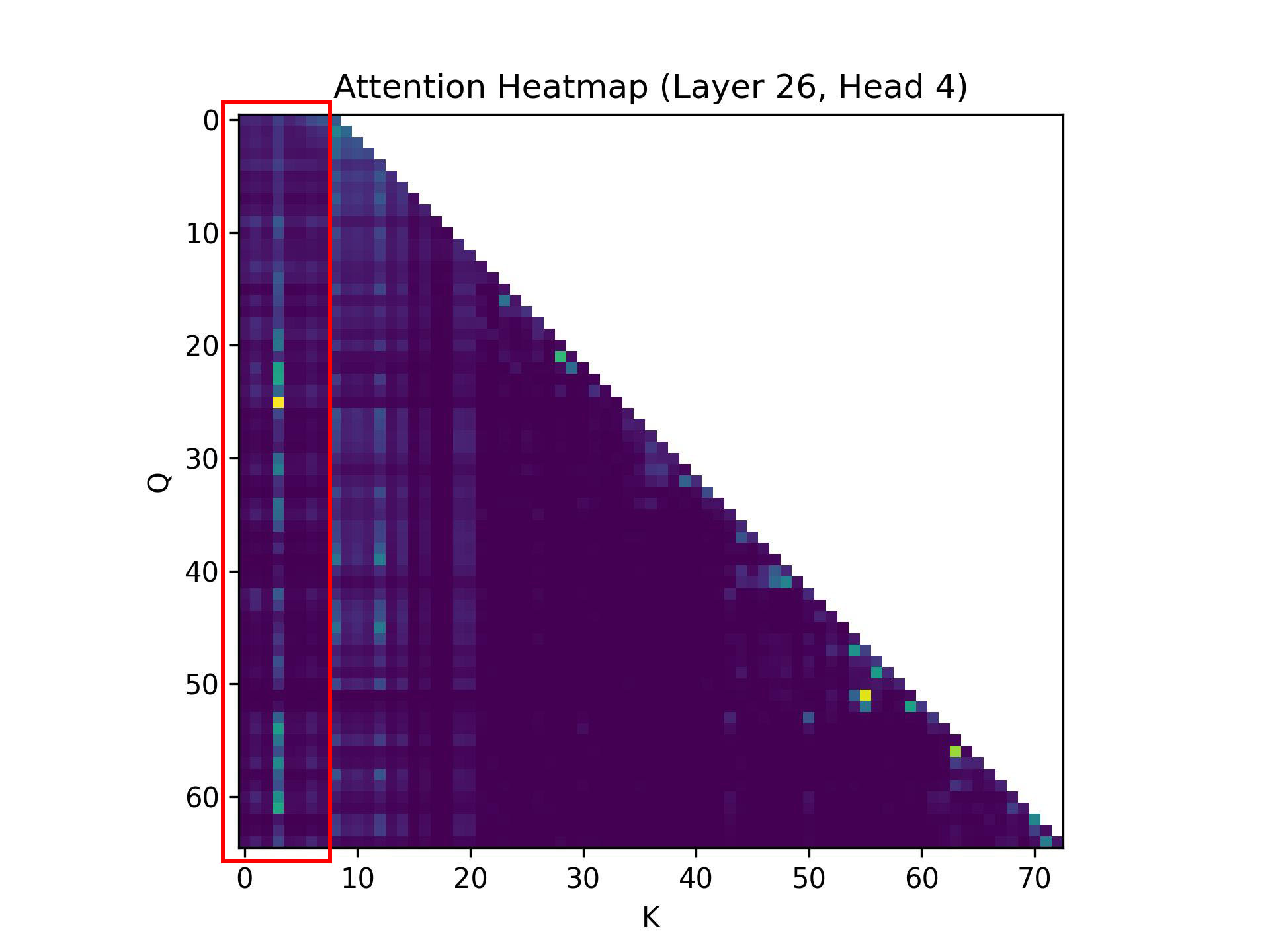} %
        \label{fig:image2}
    \end{minipage}\hfill
    \begin{minipage}[b]{0.28\textwidth}
        \centering
        \includegraphics[width=\textwidth]{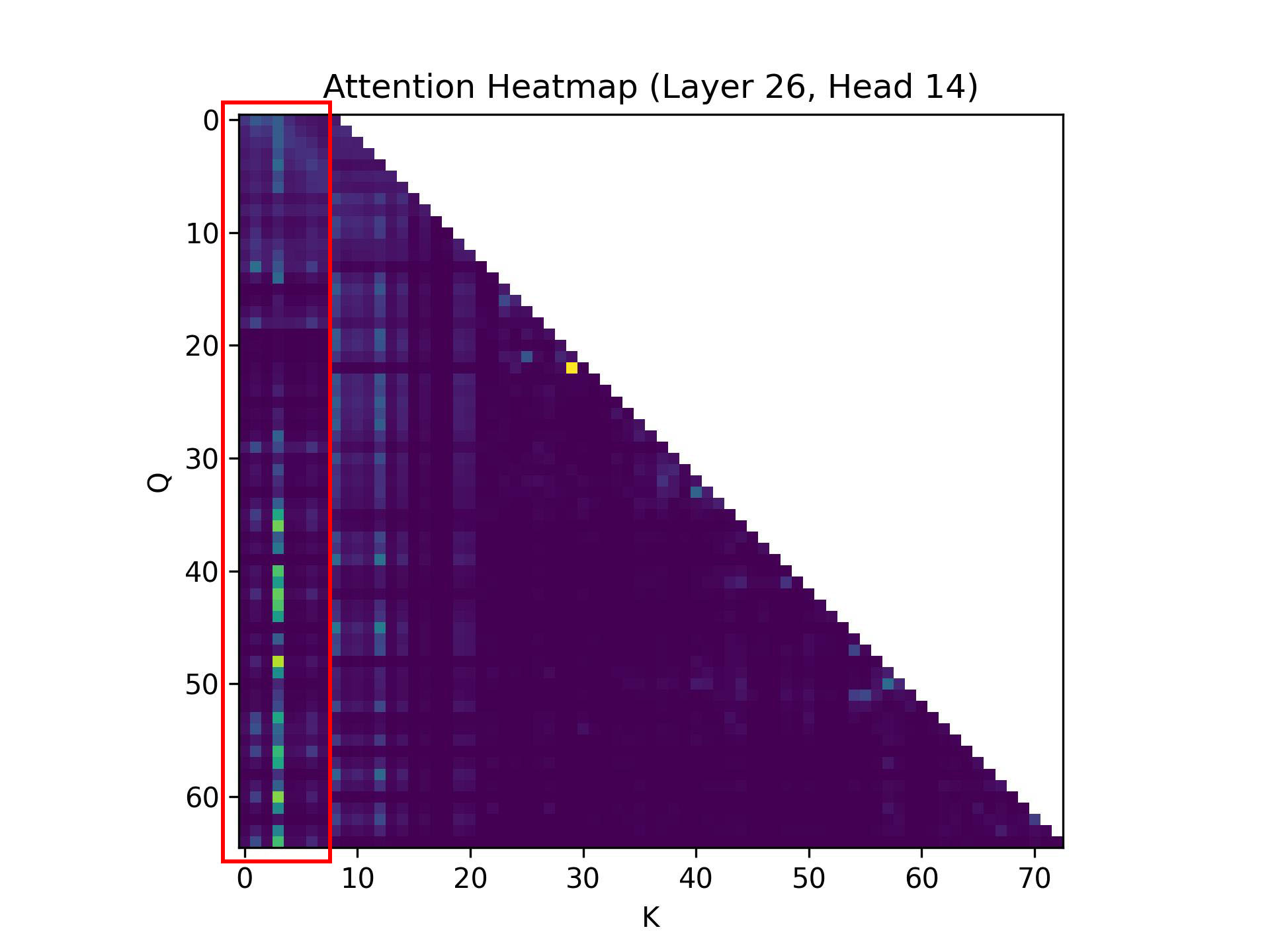} %
        \label{fig:image2}
    \end{minipage}\hfill
    \\
    \begin{minipage}[b]{0.28\textwidth} %
        \centering
        \includegraphics[width=\textwidth]{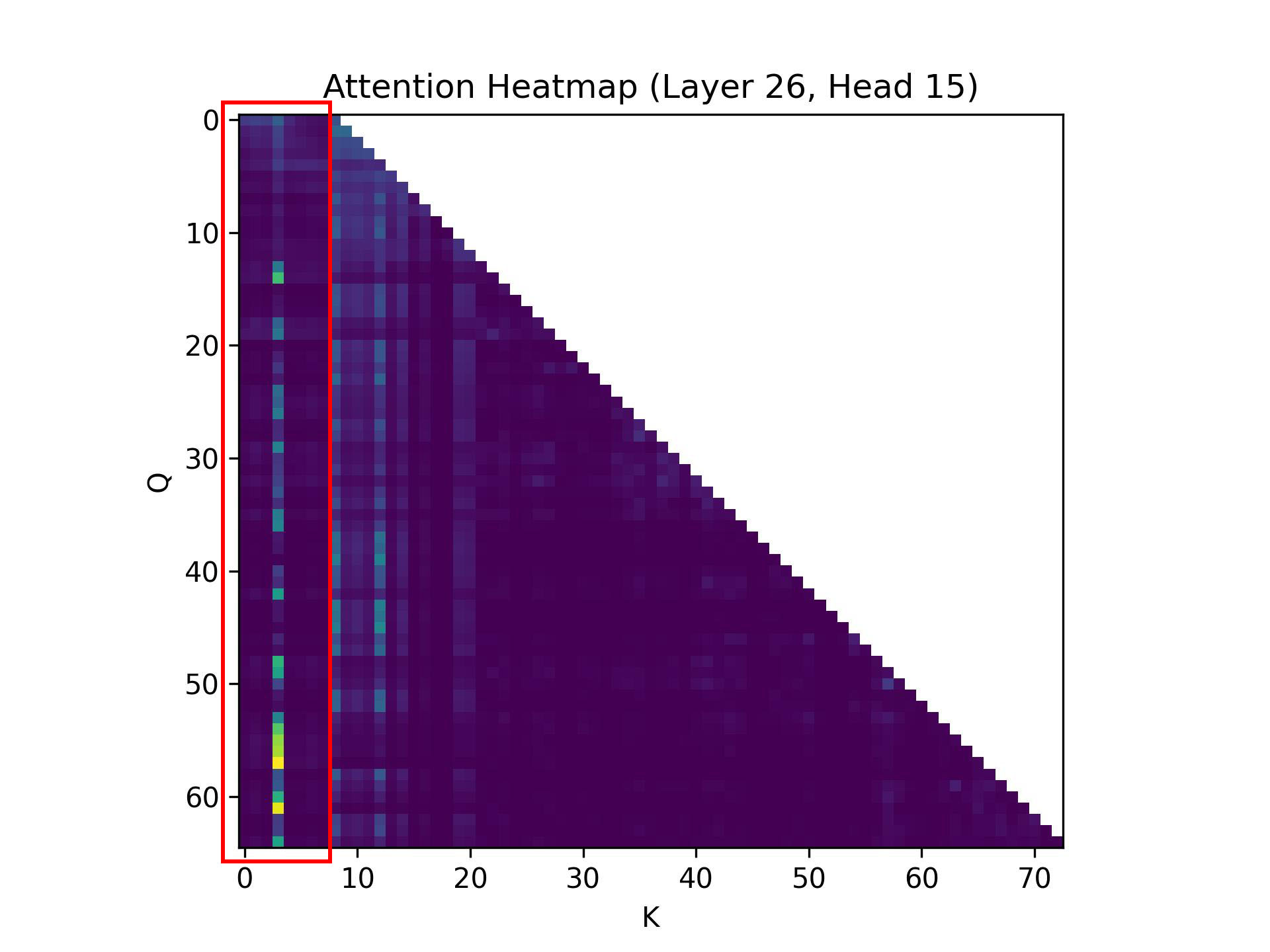} %
        \label{fig:heatmap}
    \end{minipage}\hfill
    \begin{minipage}[b]{0.28\textwidth}
        \centering
        \includegraphics[width=\textwidth]{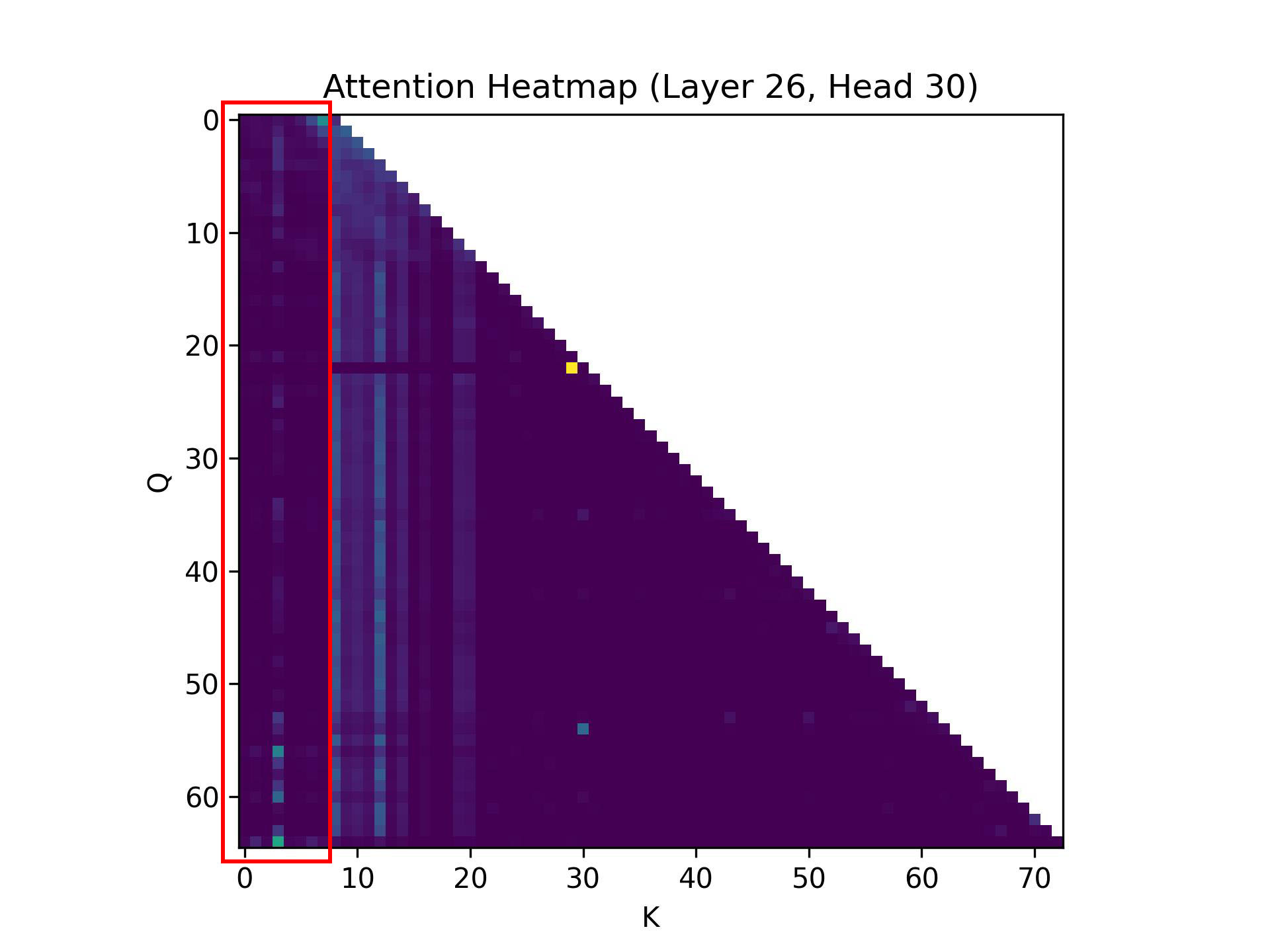} %
        \label{fig:image2}
    \end{minipage}\hfill
    \begin{minipage}[b]{0.28\textwidth} %
        \centering
        \includegraphics[width=\textwidth]{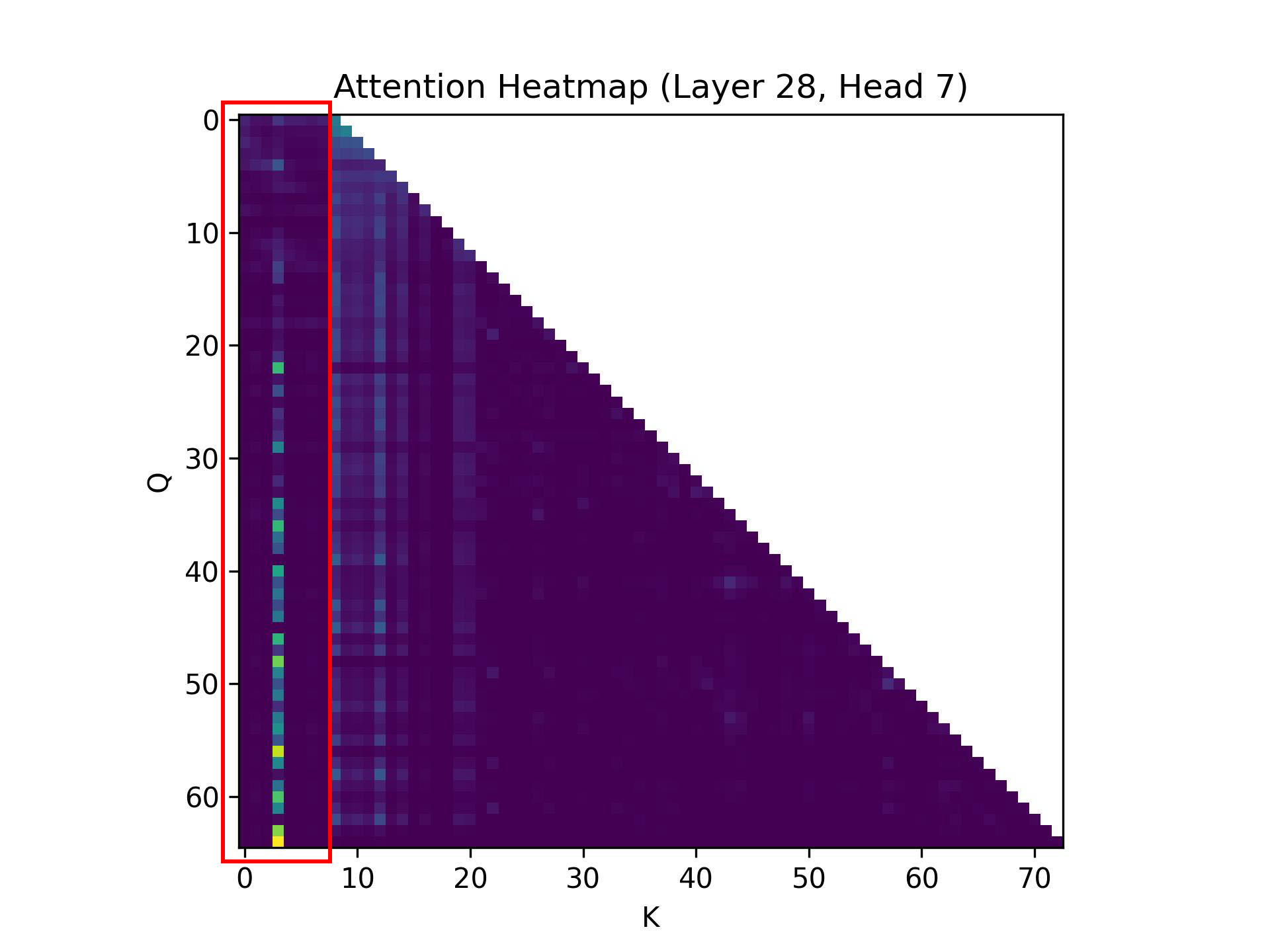} %
        \label{fig:heatmap}
    \end{minipage}\hfill
    \\
    \begin{minipage}[b]{0.28\textwidth}
        \centering
        \includegraphics[width=\textwidth]{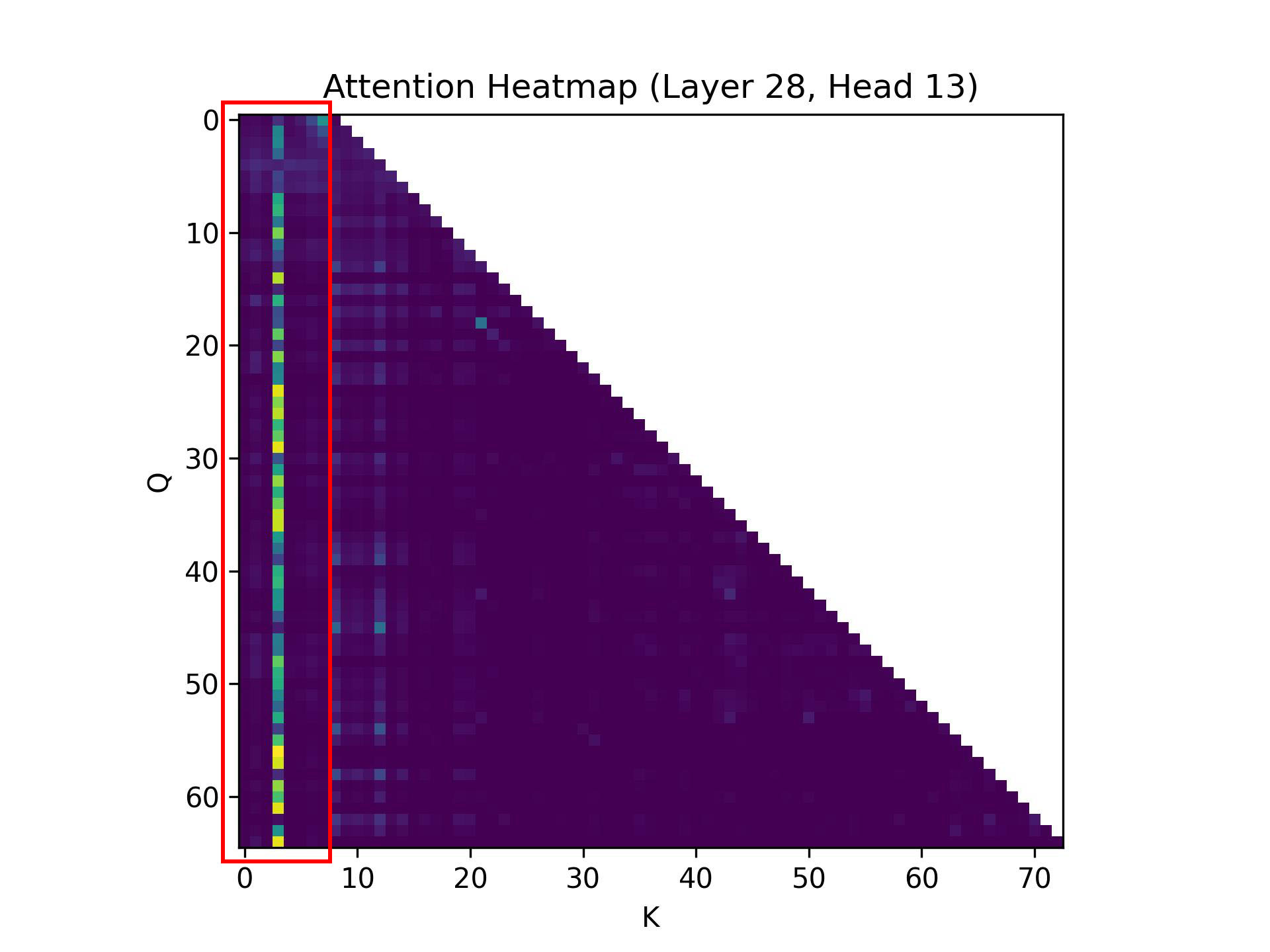} %
        \label{fig:image2}
    \end{minipage}\hfill
    \begin{minipage}[b]{0.28\textwidth} %
        \centering
        \includegraphics[width=\textwidth]{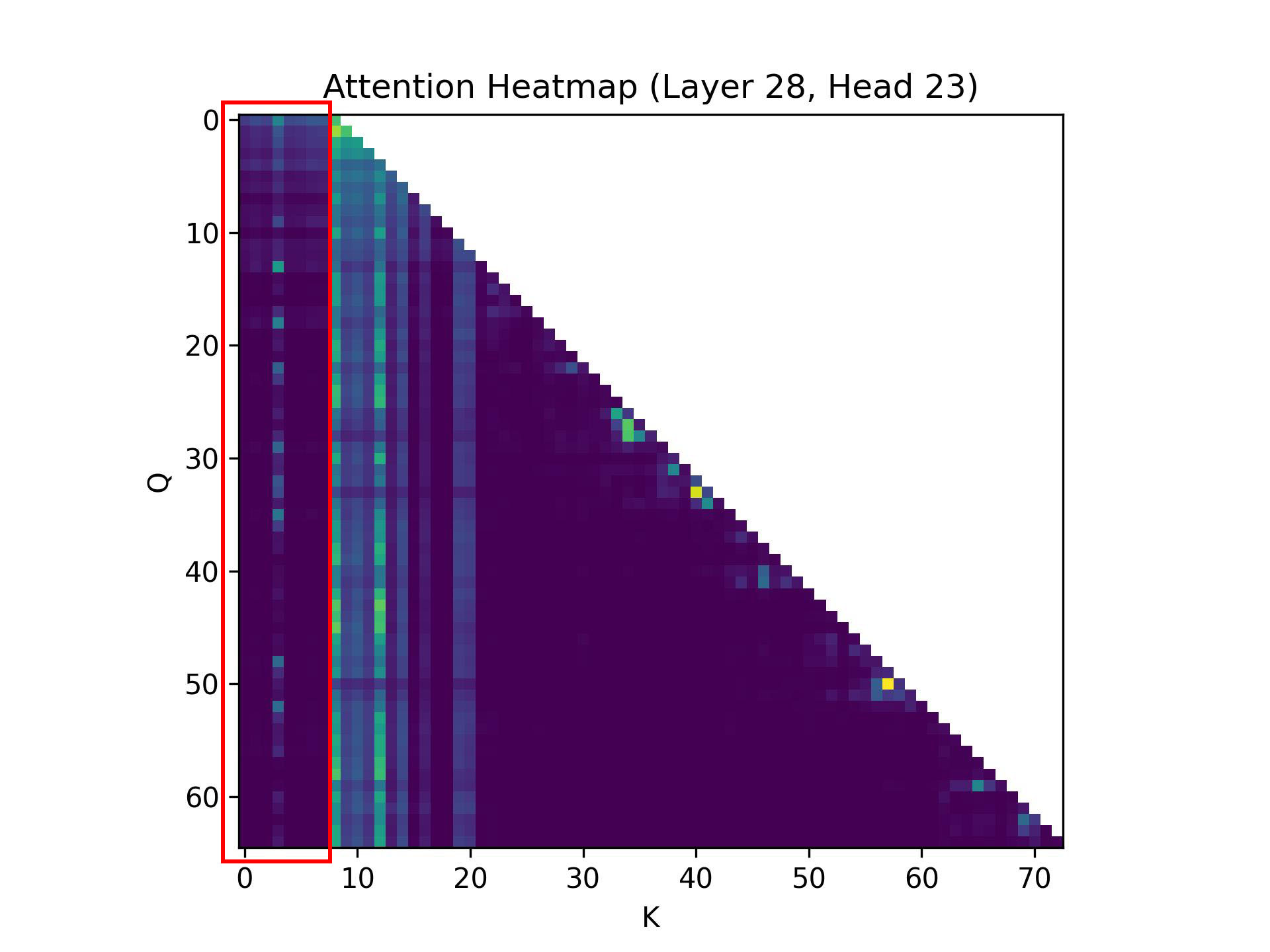} %
        \label{fig:heatmap}
    \end{minipage}\hfill
    \begin{minipage}[b]{0.28\textwidth}
        \centering
        \includegraphics[width=\textwidth]{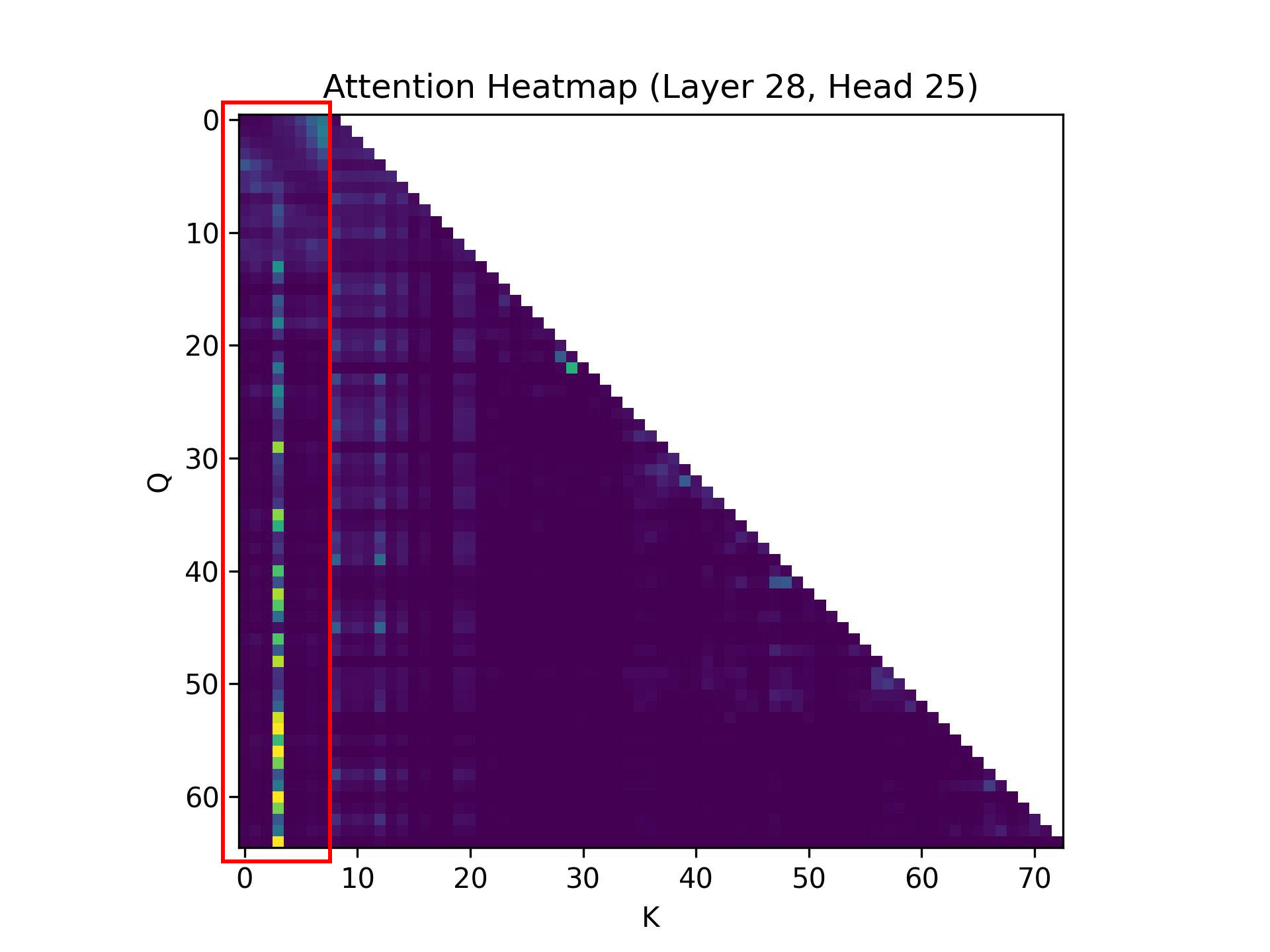} %
        \label{fig:image2}
    \end{minipage}\hfill
    \\
    \begin{minipage}[b]{0.28\textwidth} %
        \centering
        \includegraphics[width=\textwidth]{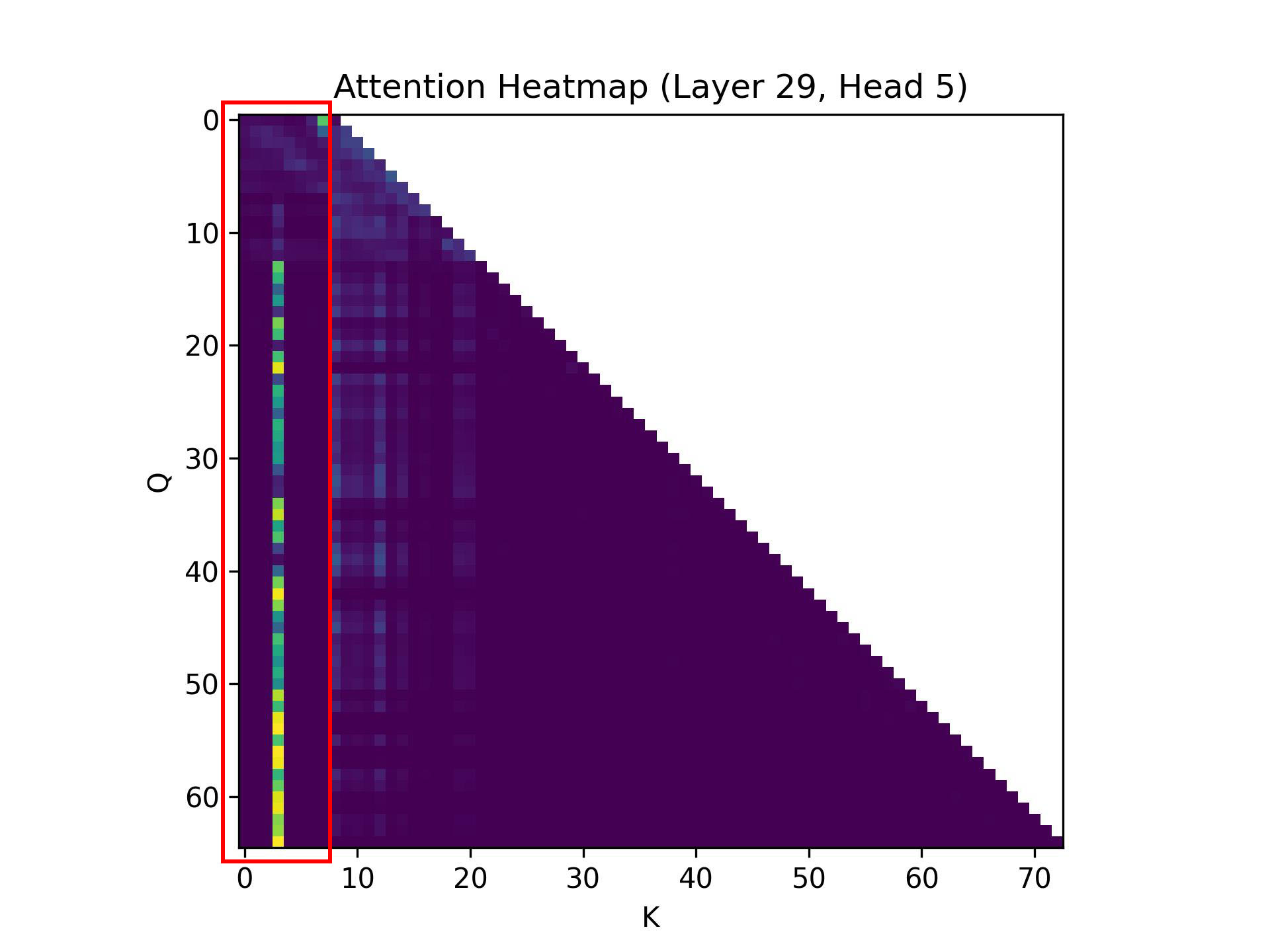} %
        \label{fig:heatmap}
    \end{minipage}\hfill
    \begin{minipage}[b]{0.28\textwidth}
        \centering
        \includegraphics[width=\textwidth]{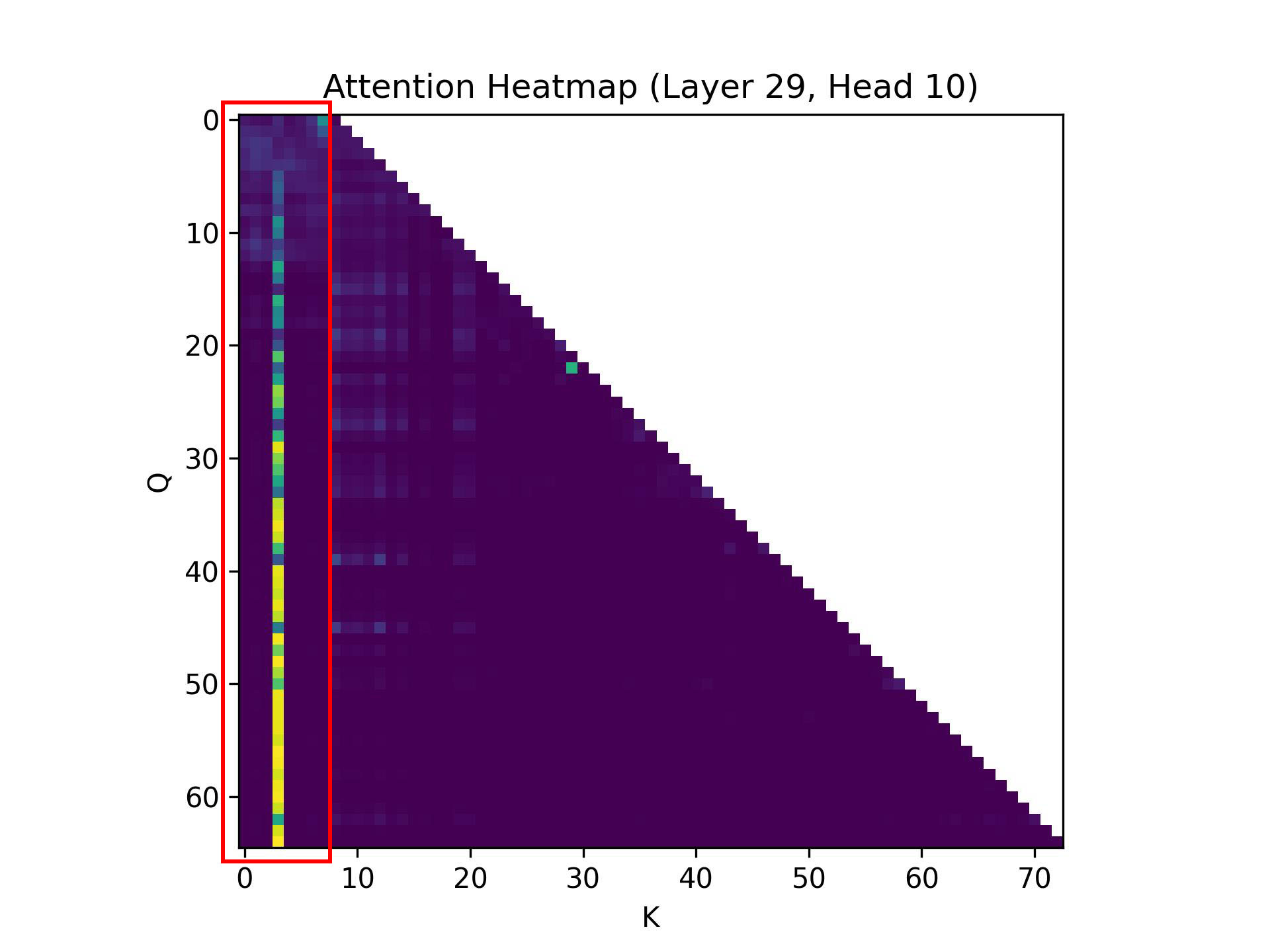} %
        \label{fig:image2}
    \end{minipage}\hfill
    \begin{minipage}[b]{0.28\textwidth} %
        \centering
        \includegraphics[width=\textwidth]{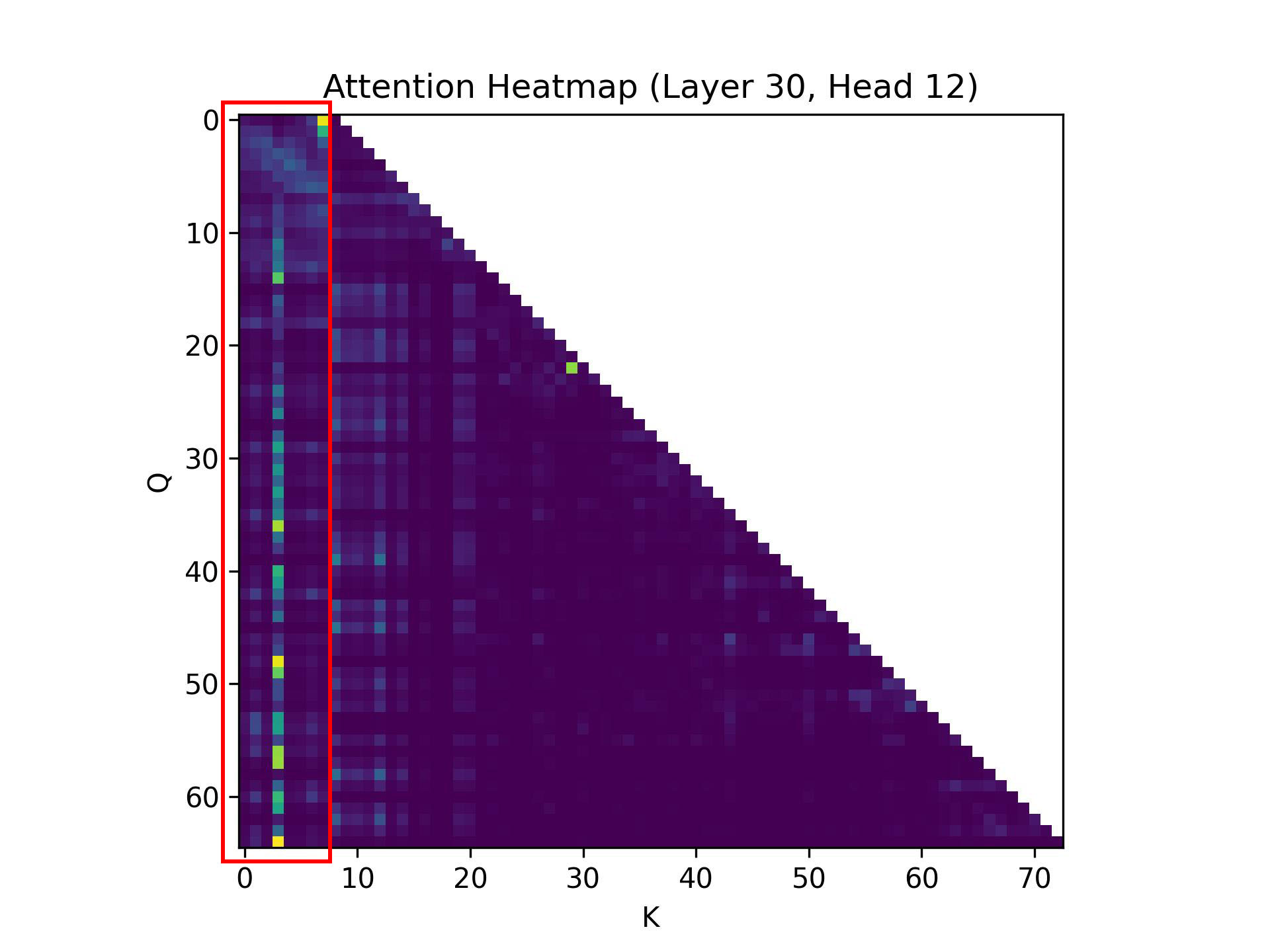} %
        \label{fig:heatmap}
    \end{minipage}\hfill
    \\
    \begin{minipage}[b]{0.28\textwidth}
        \centering
        \includegraphics[width=\textwidth]{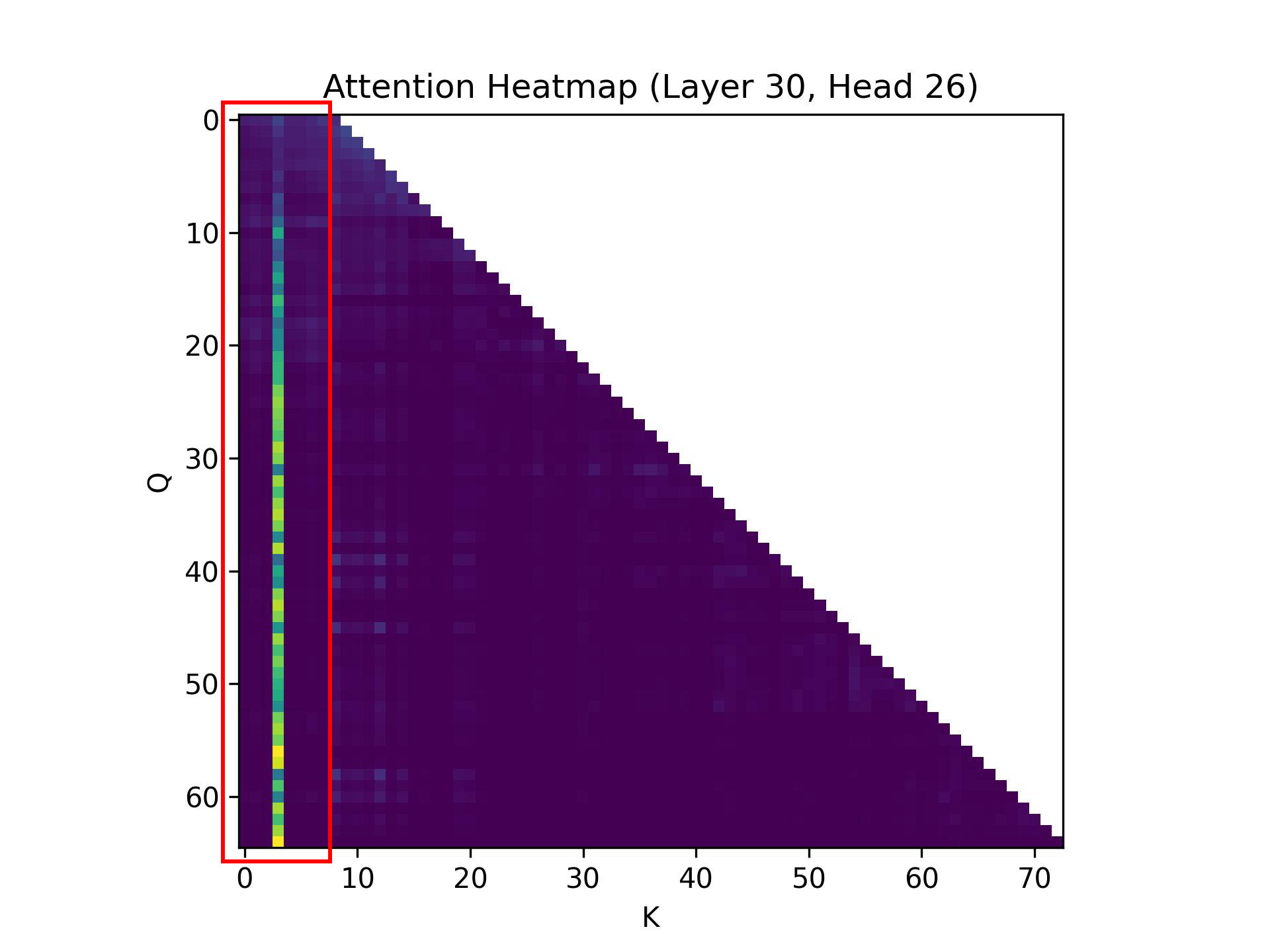} %
        \label{fig:image2}
    \end{minipage}\hfill
    \begin{minipage}[b]{0.28\textwidth}
        \centering
        \includegraphics[width=\textwidth]{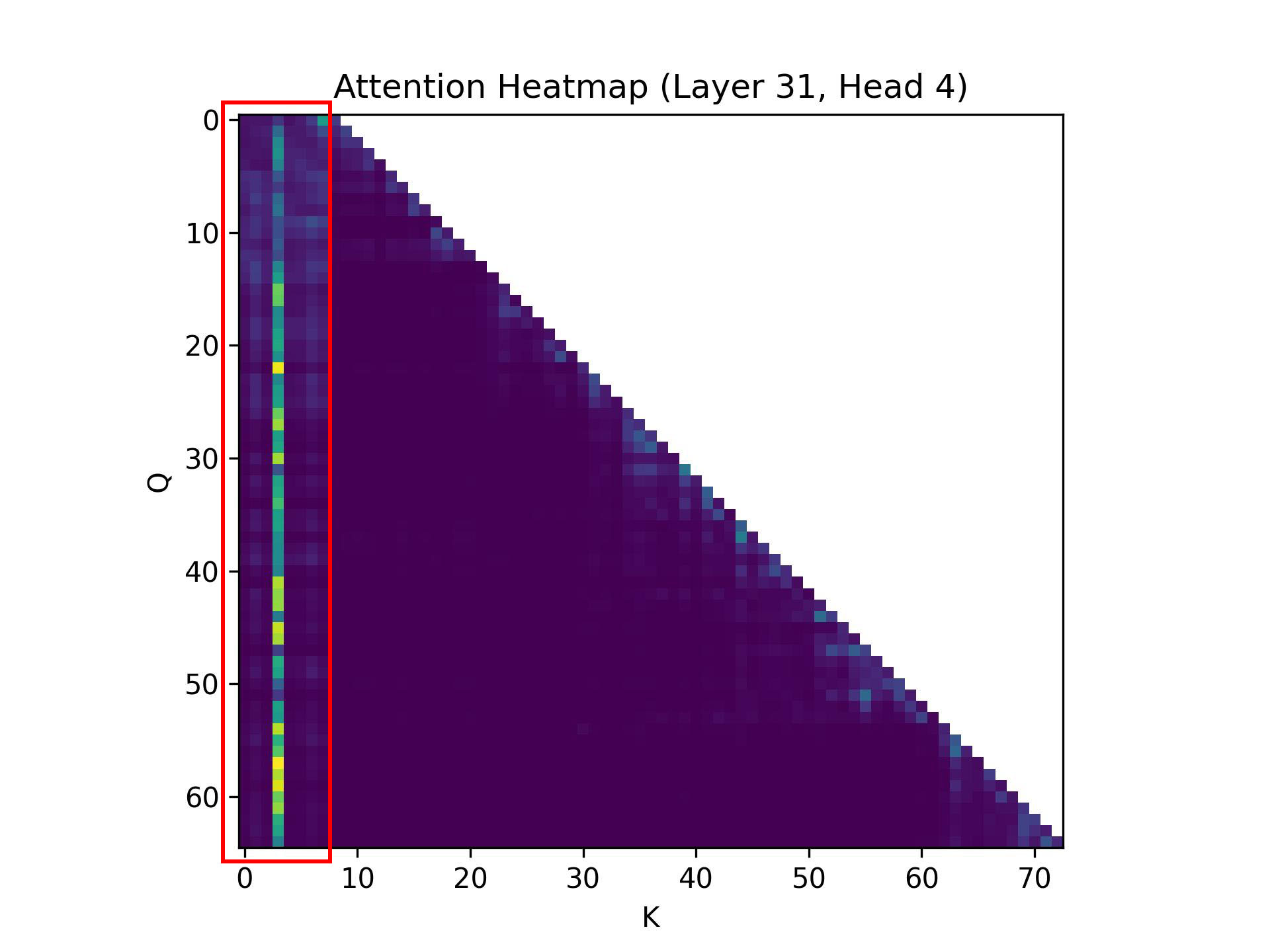} %
        \label{fig:image2}
    \end{minipage}\hfill
    \begin{minipage}[b]{0.28\textwidth}
        \centering
        \includegraphics[width=\textwidth]{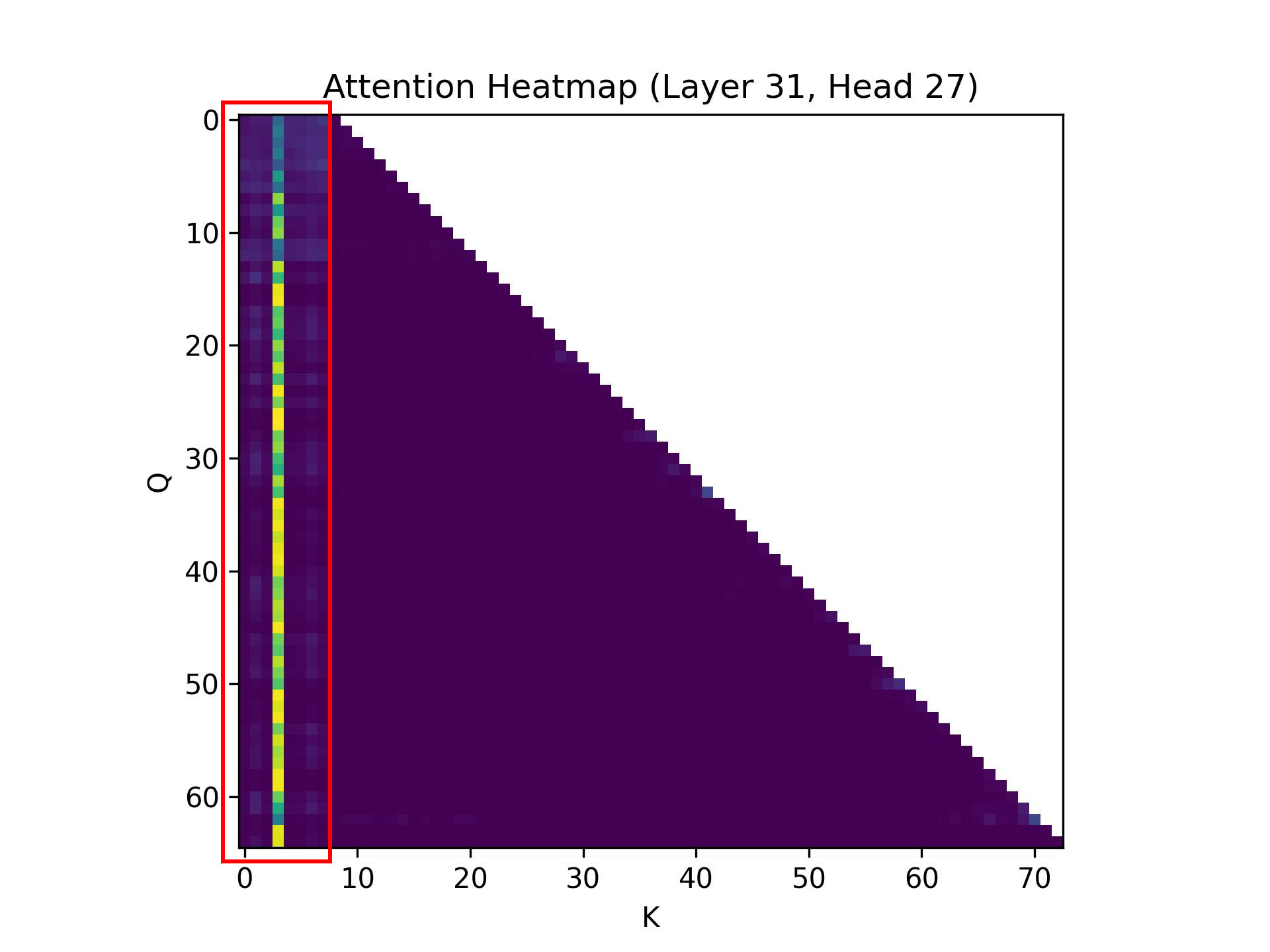} %
        \label{fig:image2}
    \end{minipage}\hfill
    \caption{Attention Heamap of Passkey Retrieval task. The first 8 columns, marked by red rectangule lines, represent the attention weights corresponding to 8 candidate tokens. Since only one chunk contains the important passkey information while the others are merely noises, we observe that only a single candidate token receives high attention score (with a single column highlighted). This suggests that FocusLLM can effectively extract important information while discarding irrelevant texts.}
    \label{Attention Heamap for Passkey Retrieval task}
\end{figure*}

\begin{figure*}[ht]
    \centering
    \begin{minipage}[b]{0.28\textwidth} %
        \centering
        \includegraphics[width=\textwidth]{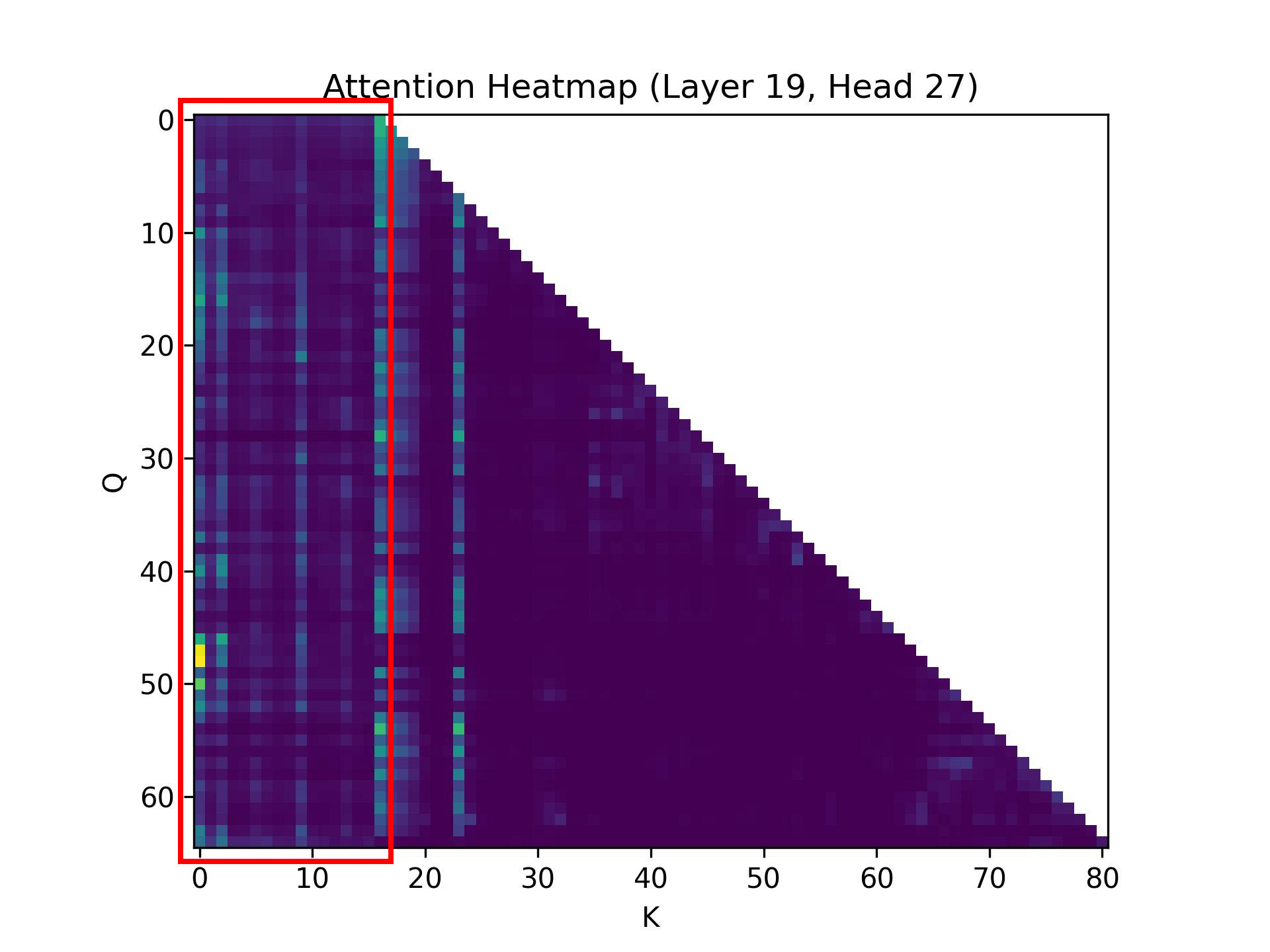} %
        \label{fig:heatmap}
    \end{minipage}\hfill
    \begin{minipage}[b]{0.28\textwidth} %
        \centering
        \includegraphics[width=\textwidth]{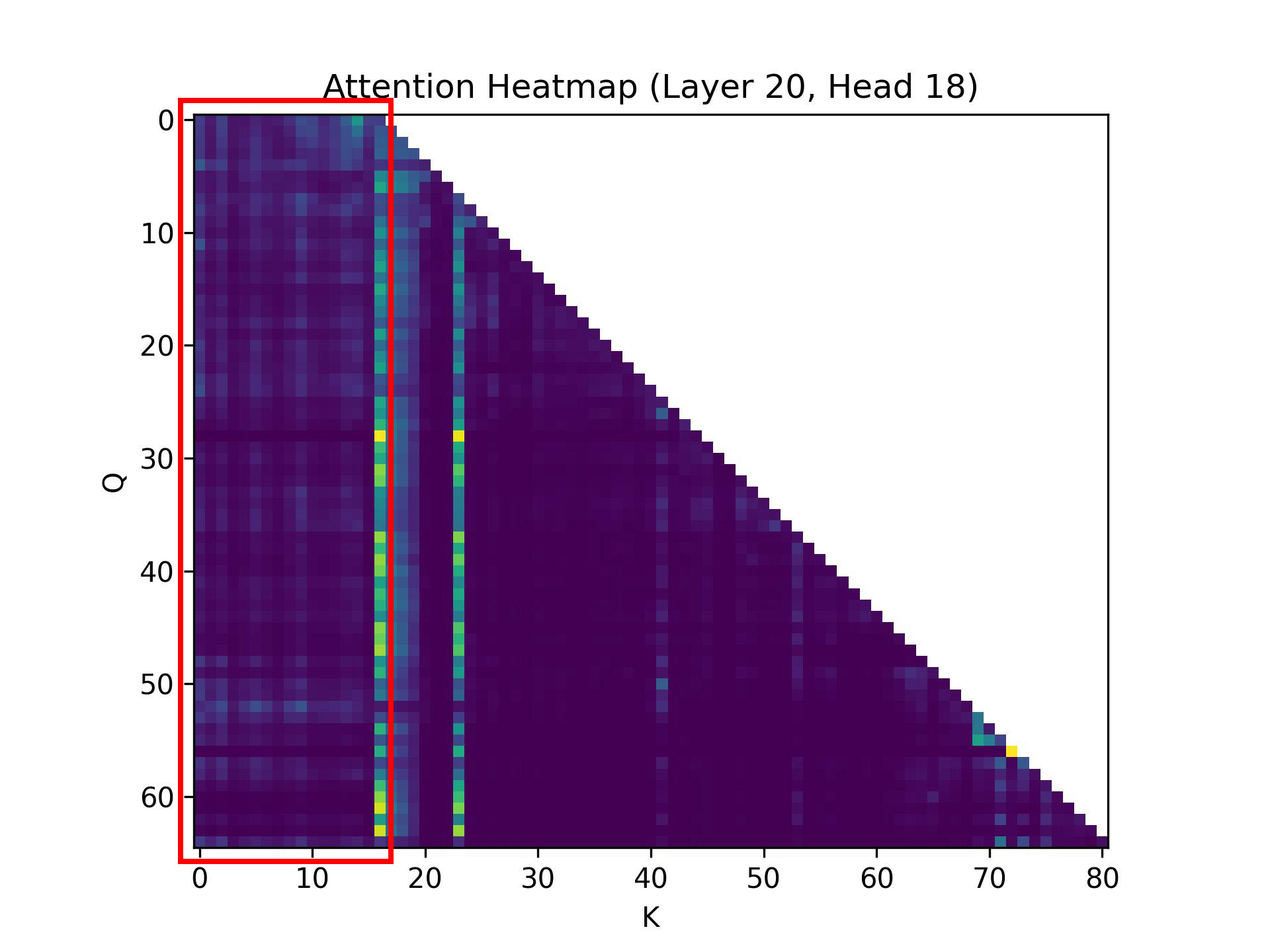} %
        \label{fig:heatmap}
    \end{minipage}\hfill
    \begin{minipage}[b]{0.28\textwidth}
        \centering
        \includegraphics[width=\textwidth]{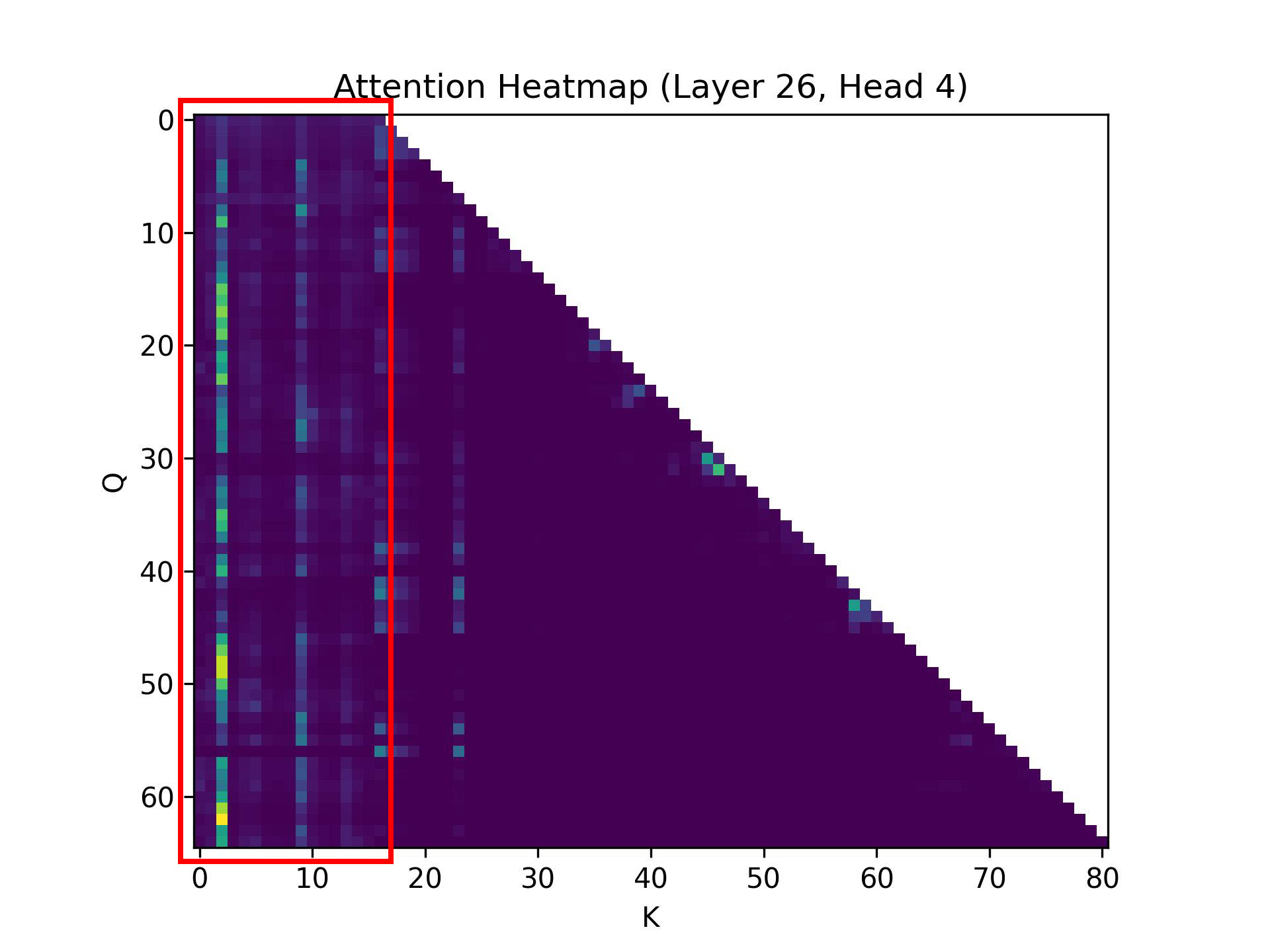} %
        \label{fig:image2}
    \end{minipage}\hfill
    \\
    \begin{minipage}[b]{0.28\textwidth}
        \centering
        \includegraphics[width=\textwidth]{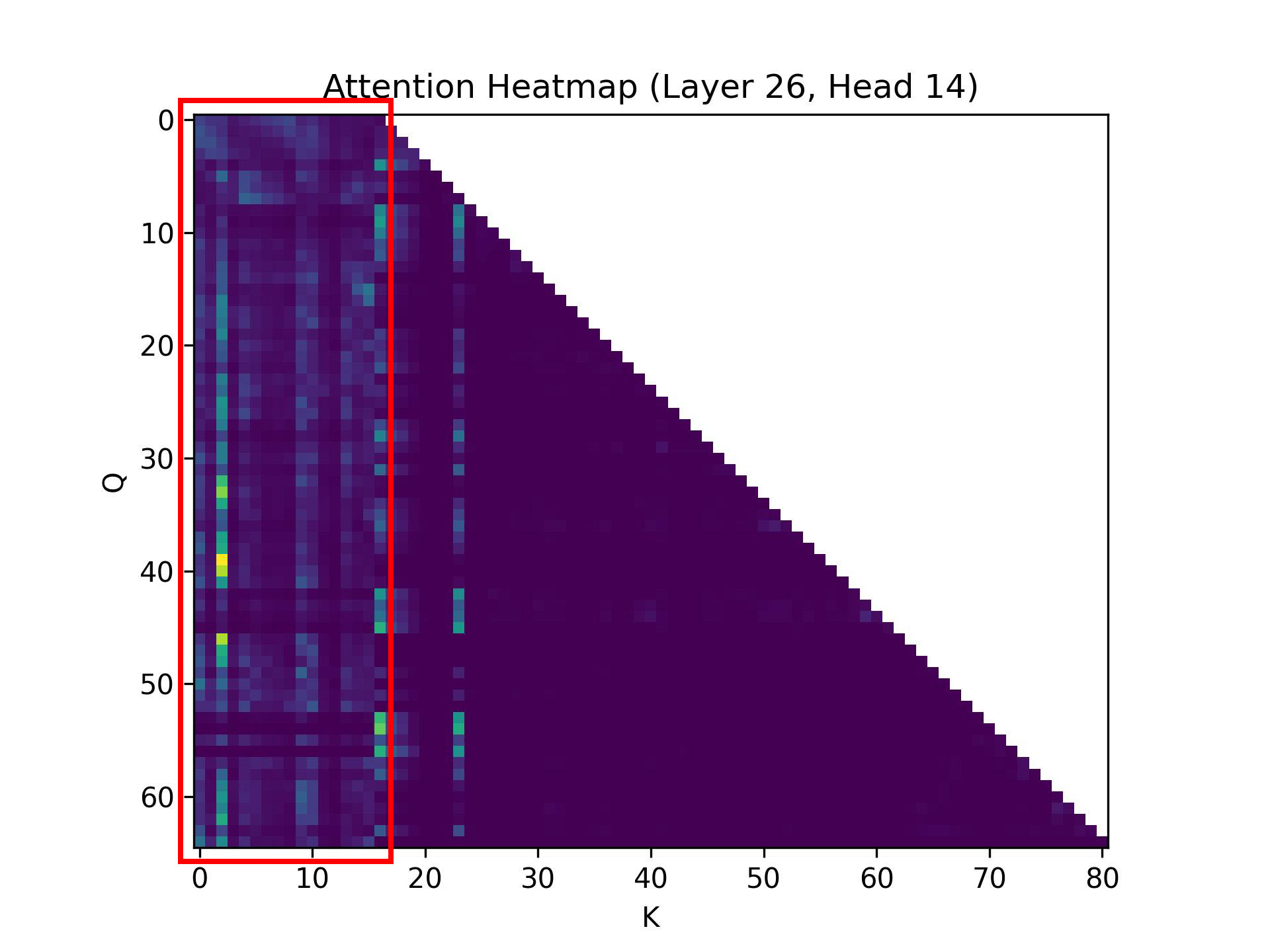} %
        \label{fig:image2}
    \end{minipage}\hfill
    \begin{minipage}[b]{0.28\textwidth} %
        \centering
        \includegraphics[width=\textwidth]{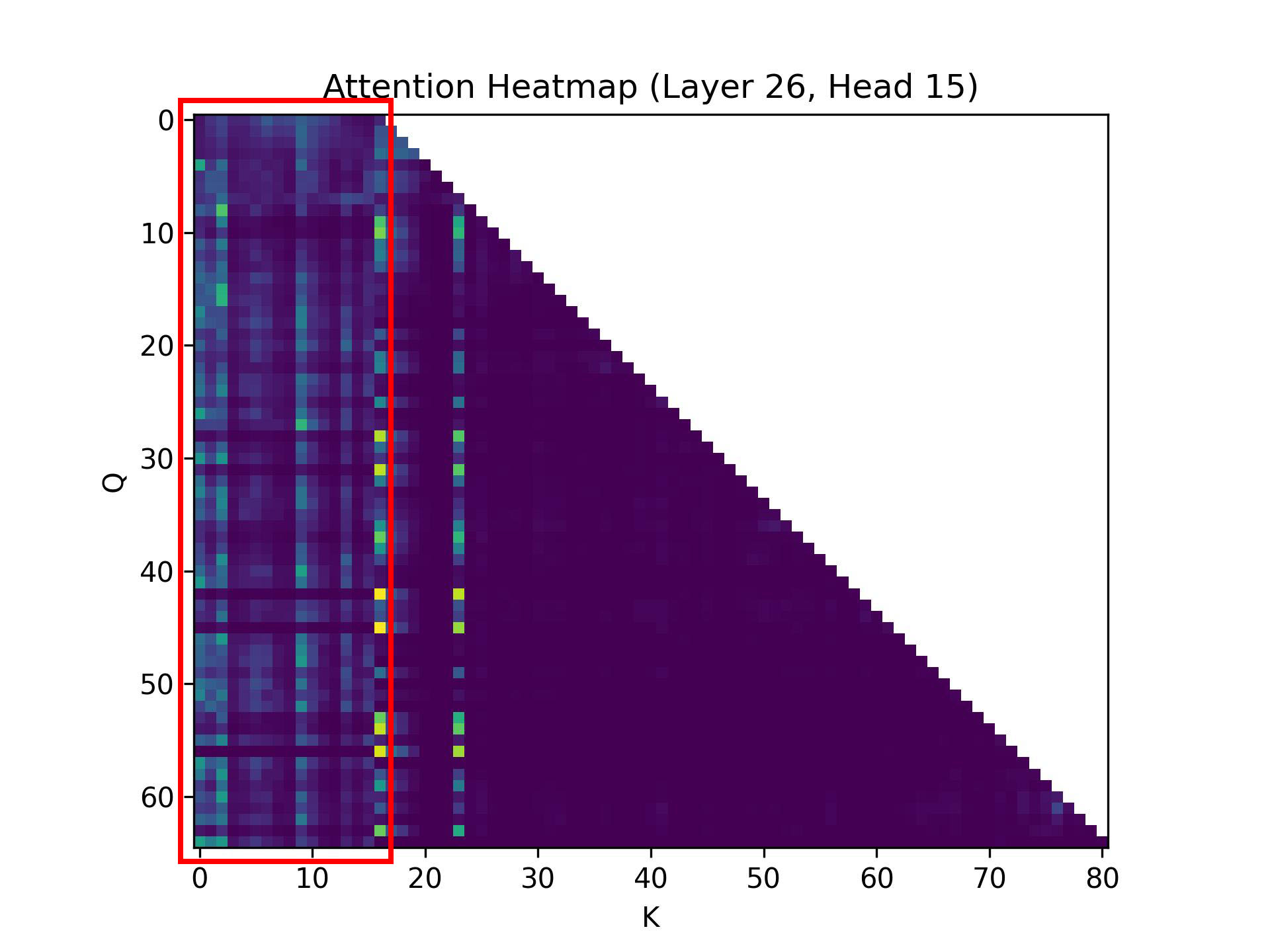} %
        \label{fig:heatmap}
    \end{minipage}\hfill
    \begin{minipage}[b]{0.28\textwidth} %
        \centering
        \includegraphics[width=\textwidth]{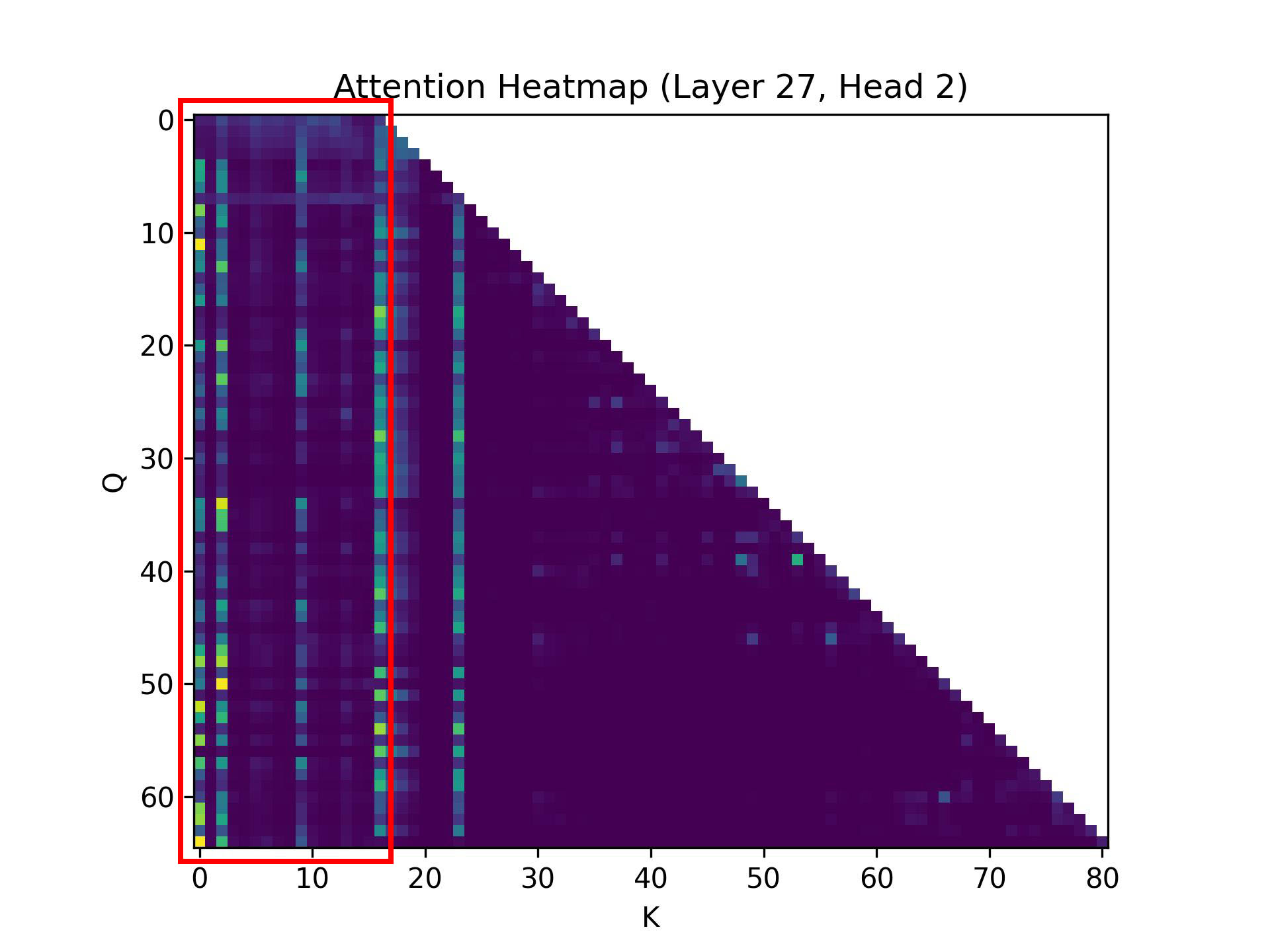} %
        \label{fig:heatmap}
    \end{minipage}\hfill
    \\
    \begin{minipage}[b]{0.28\textwidth} %
        \centering
        \includegraphics[width=\textwidth]{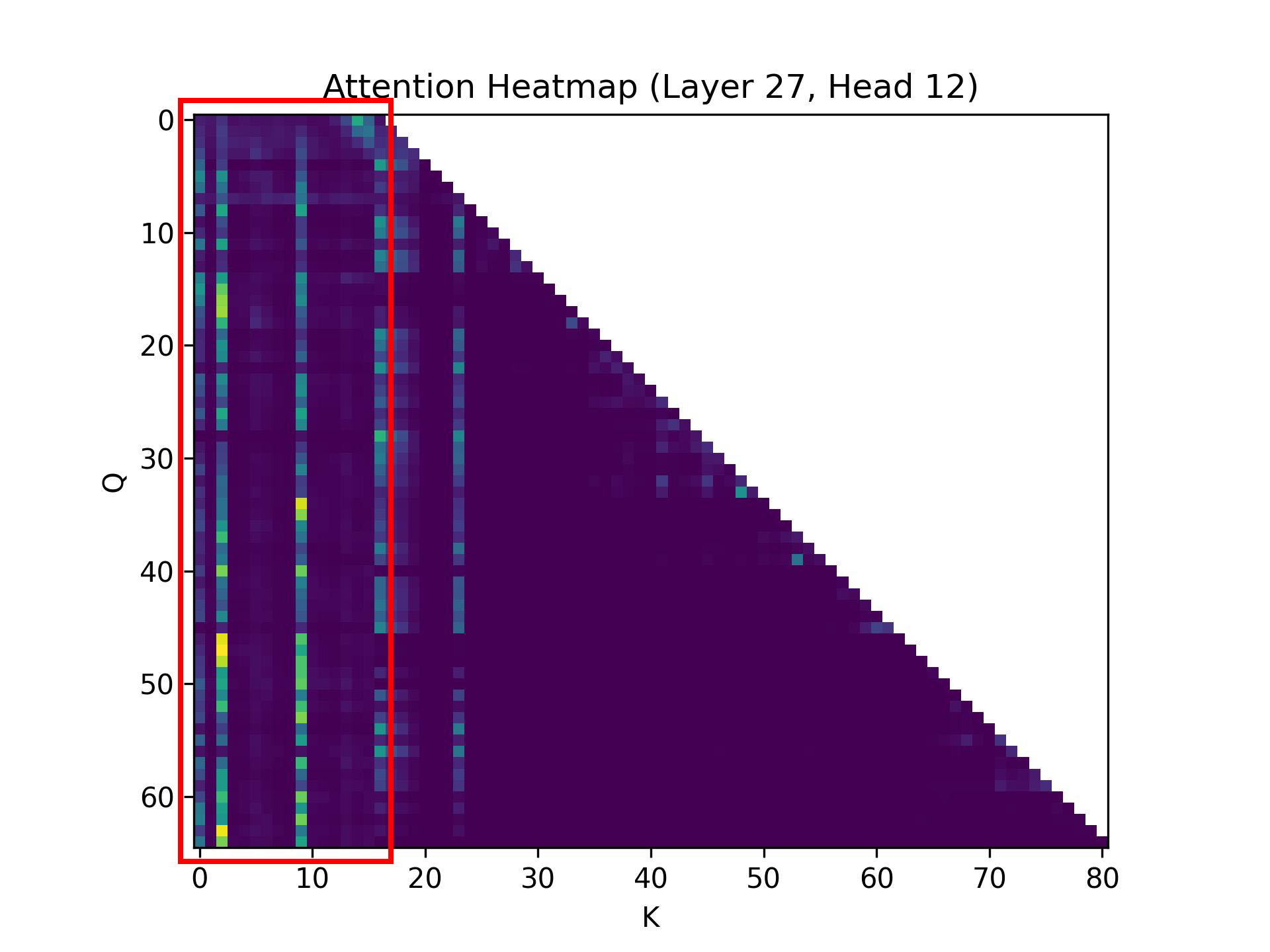} %
        \label{fig:heatmap}
    \end{minipage}\hfill
    \begin{minipage}[b]{0.28\textwidth}
        \centering
        \includegraphics[width=\textwidth]{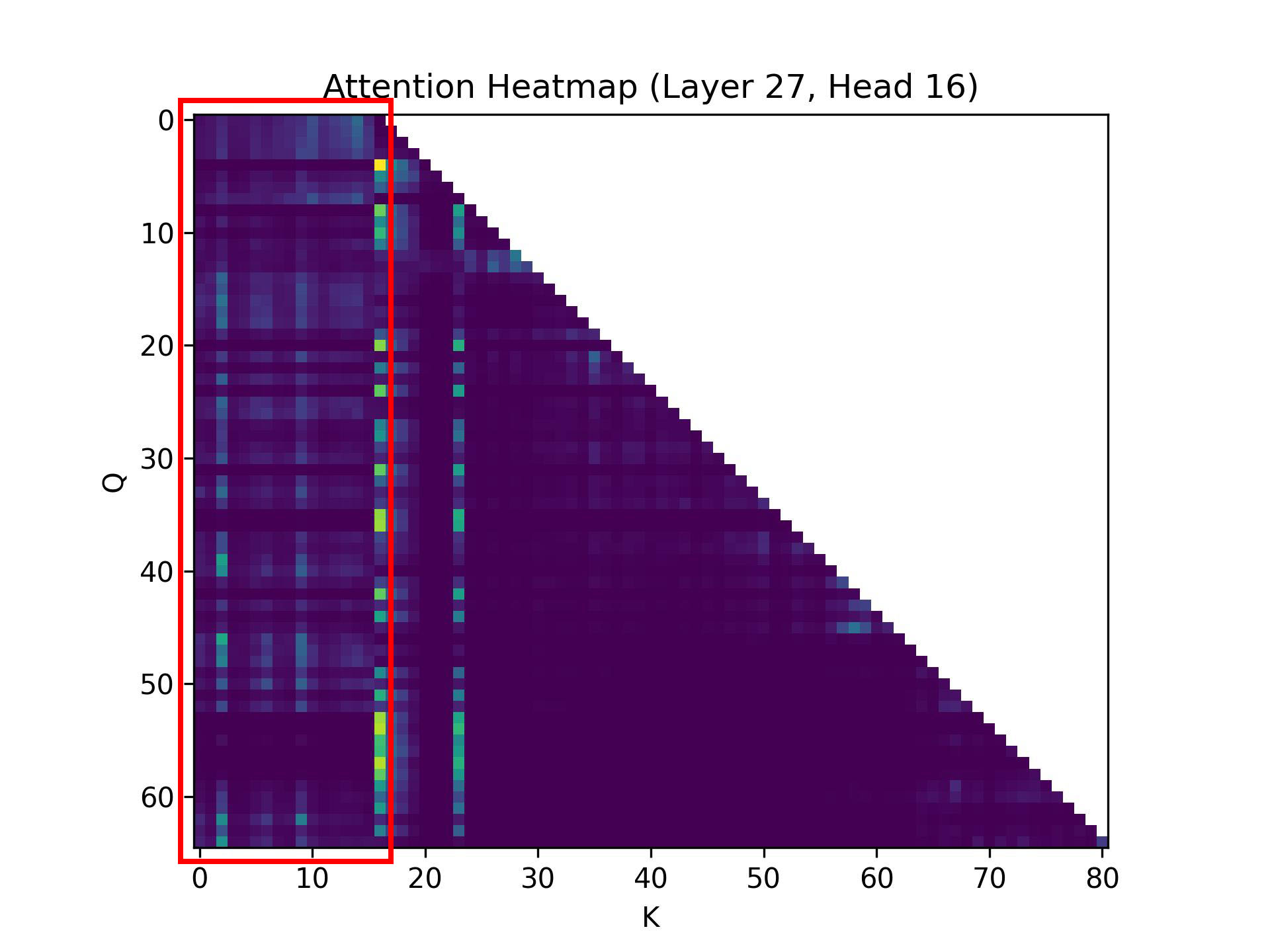} %
        \label{fig:image2}
    \end{minipage}\hfill
    \begin{minipage}[b]{0.28\textwidth} %
        \centering
        \includegraphics[width=\textwidth]{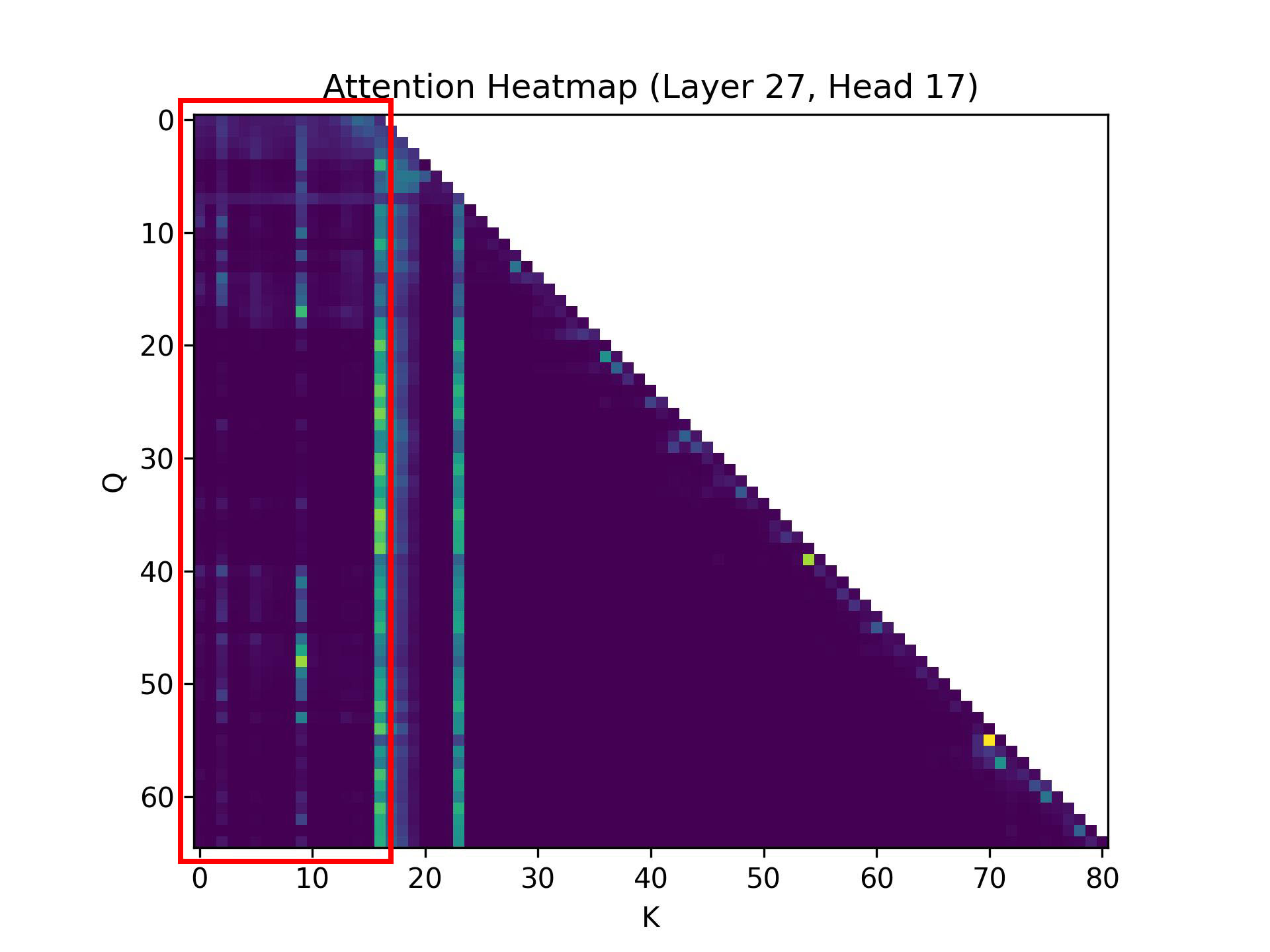} %
        \label{fig:heatmap}
    \end{minipage}\hfill
    \\
    \begin{minipage}[b]{0.28\textwidth}
        \centering
        \includegraphics[width=\textwidth]{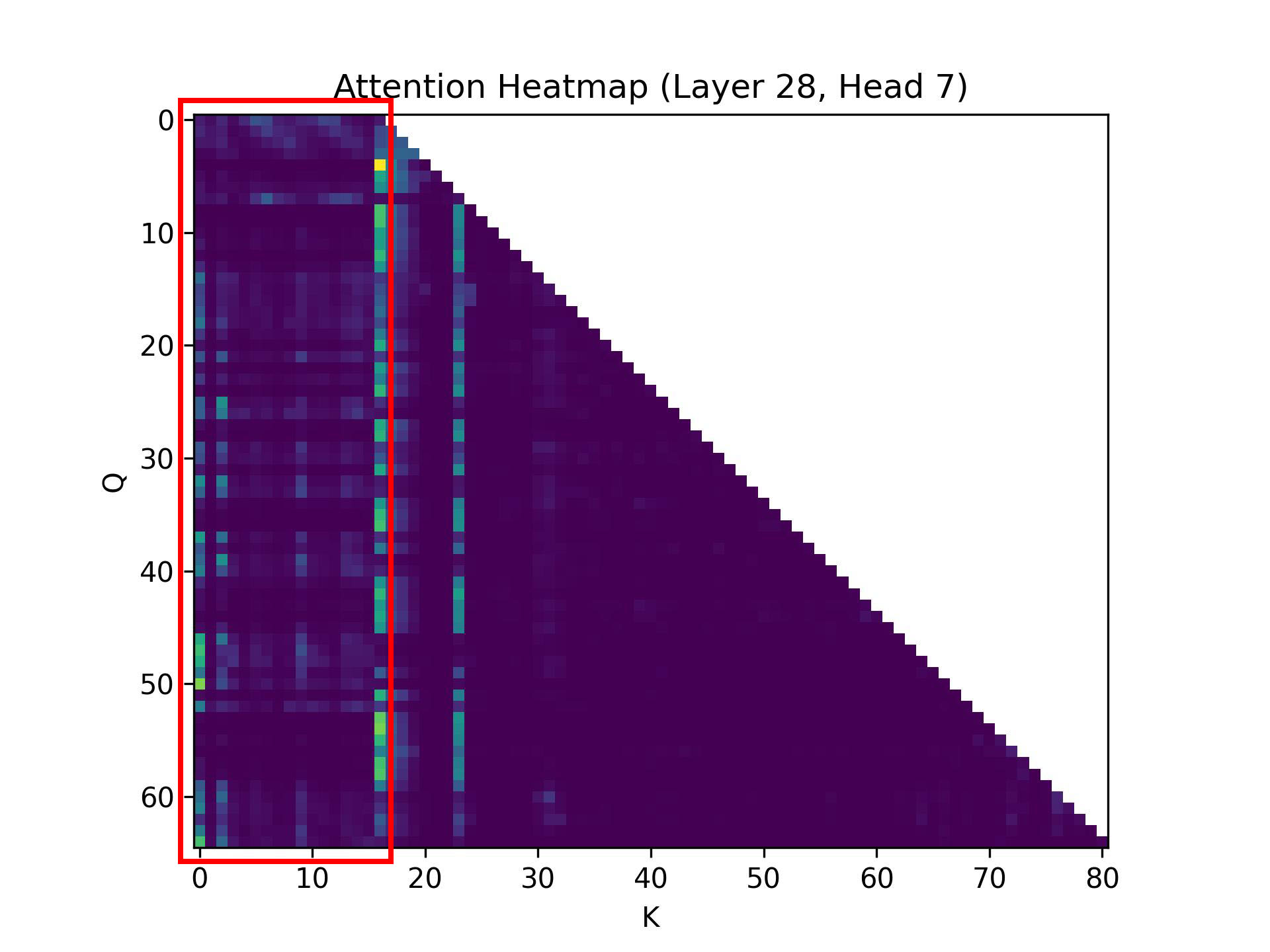} %
        \label{fig:image2}
    \end{minipage}\hfill
    \begin{minipage}[b]{0.28\textwidth} %
        \centering
        \includegraphics[width=\textwidth]{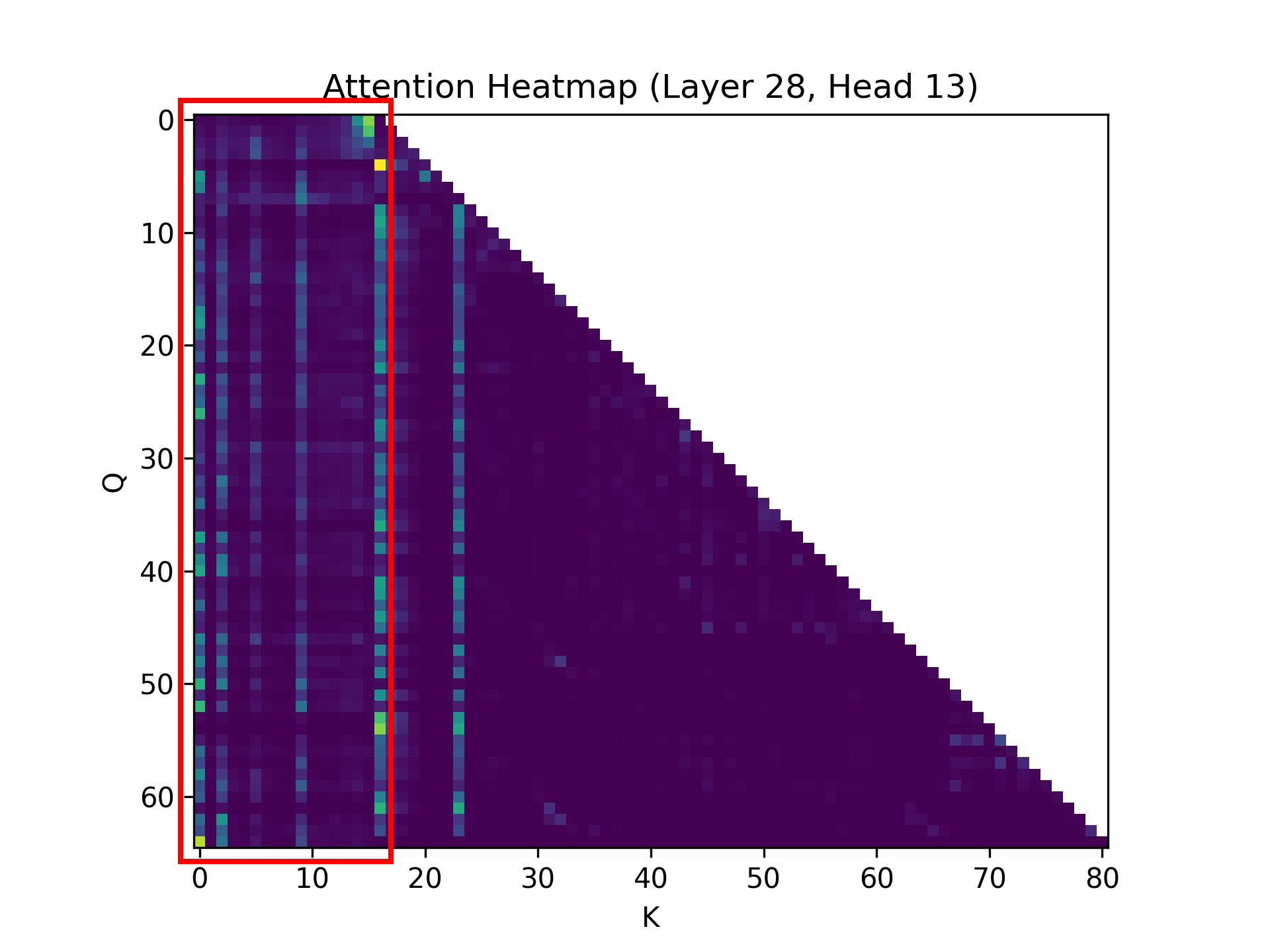} %
        \label{fig:heatmap}
    \end{minipage}\hfill
    \begin{minipage}[b]{0.28\textwidth}
        \centering
        \includegraphics[width=\textwidth]{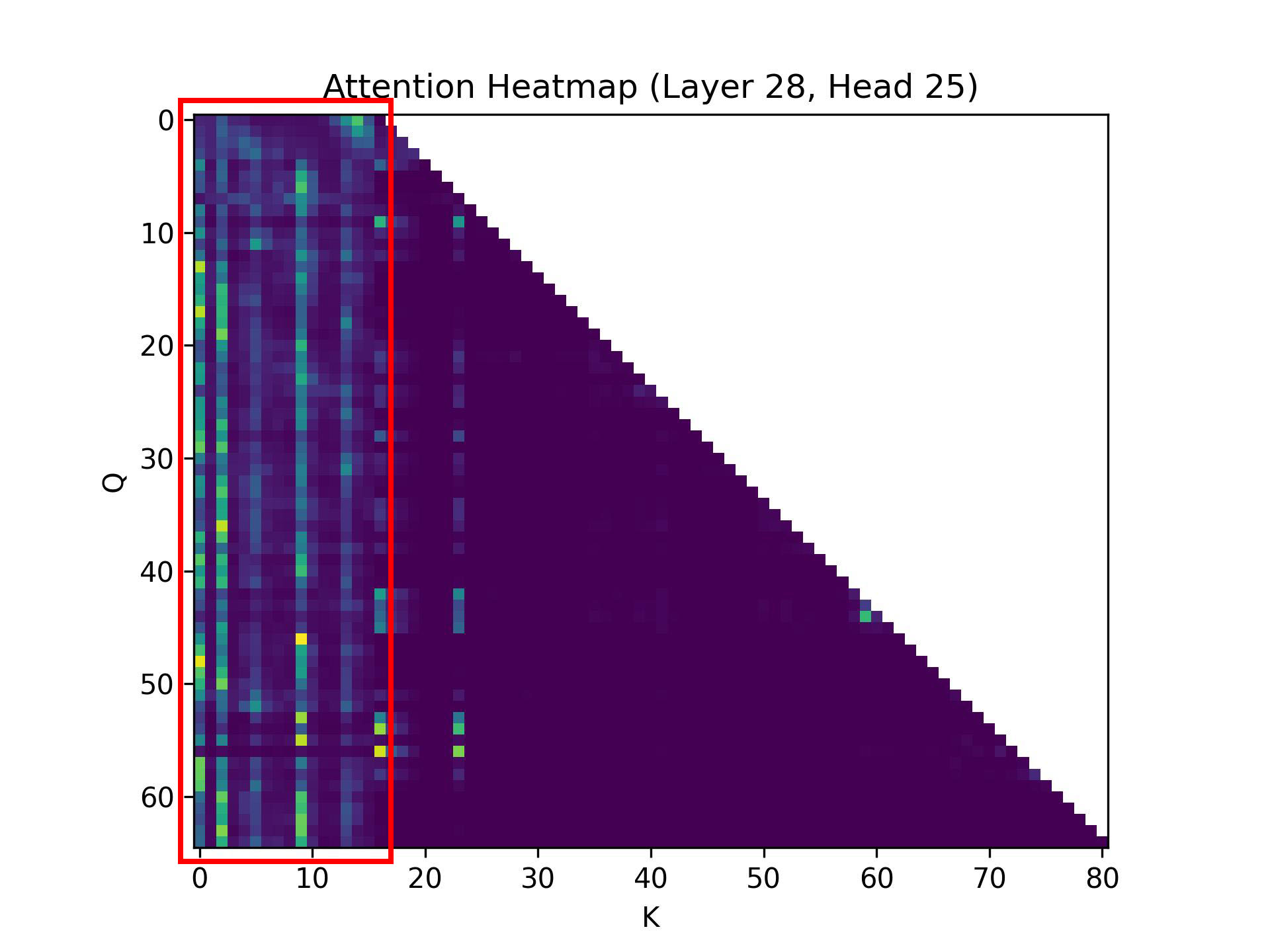} %
        \label{fig:image2}
    \end{minipage}\hfill
    \\
    \begin{minipage}[b]{0.28\textwidth}
        \centering
        \includegraphics[width=\textwidth]{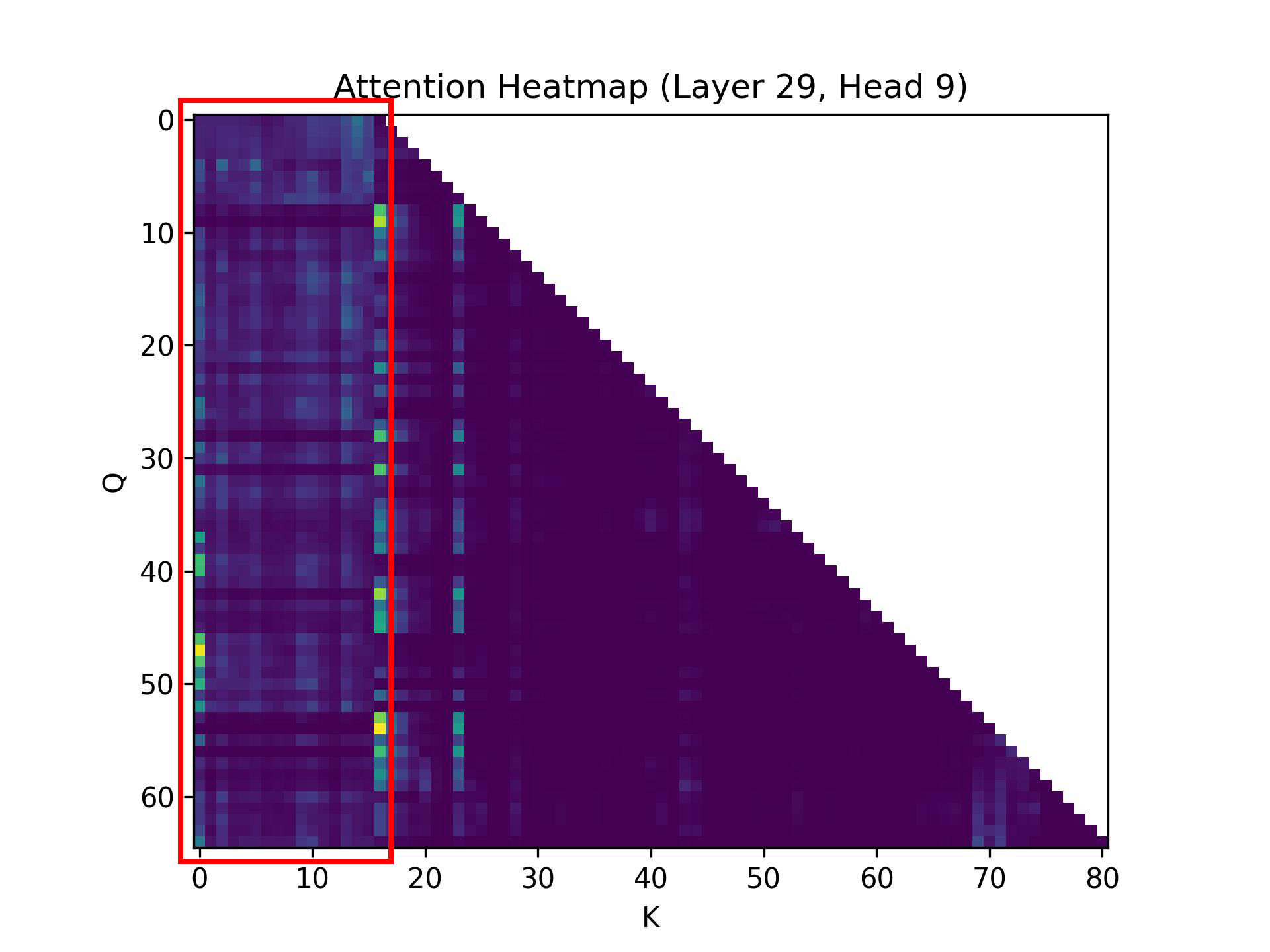} %
        \label{fig:image2}
    \end{minipage}\hfill
    \begin{minipage}[b]{0.28\textwidth}
        \centering
        \includegraphics[width=\textwidth]{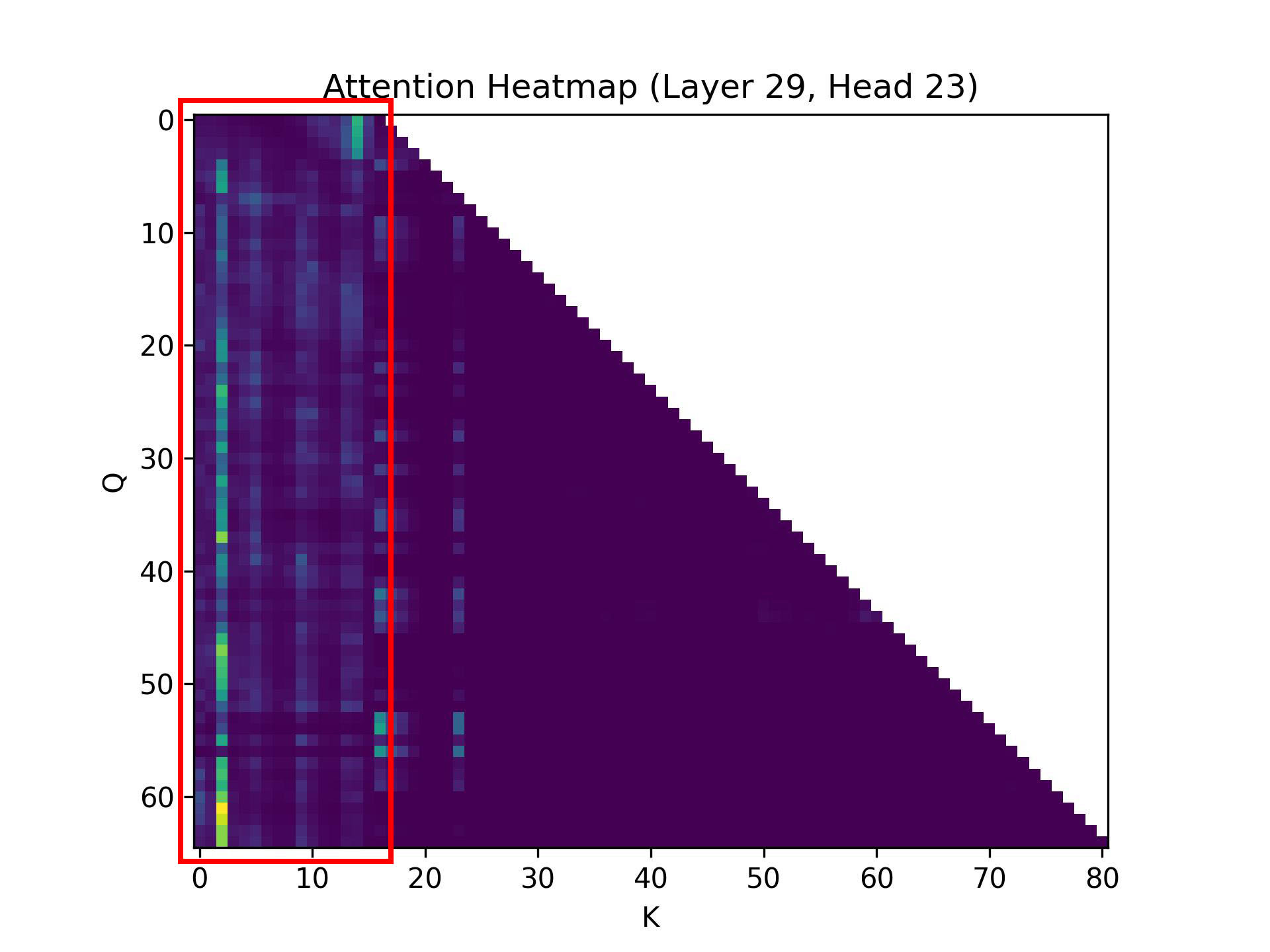} %
        \label{fig:image2}
    \end{minipage}\hfill
    \begin{minipage}[b]{0.28\textwidth}
        \centering
        \includegraphics[width=\textwidth]{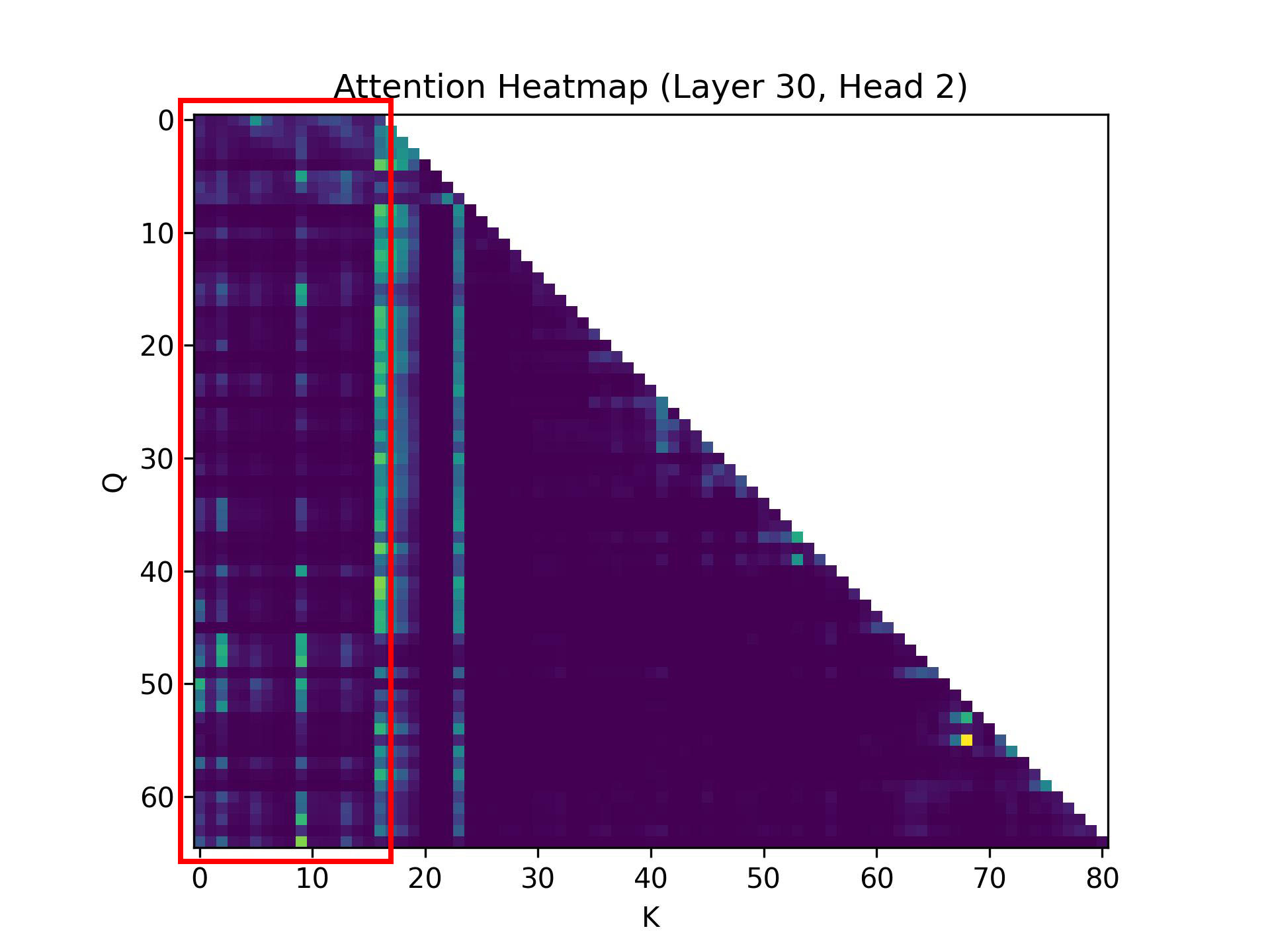} %
        \label{fig:image2}
    \end{minipage}\hfill
    \\
    \begin{minipage}[b]{0.28\textwidth}
        \centering
        \includegraphics[width=\textwidth]{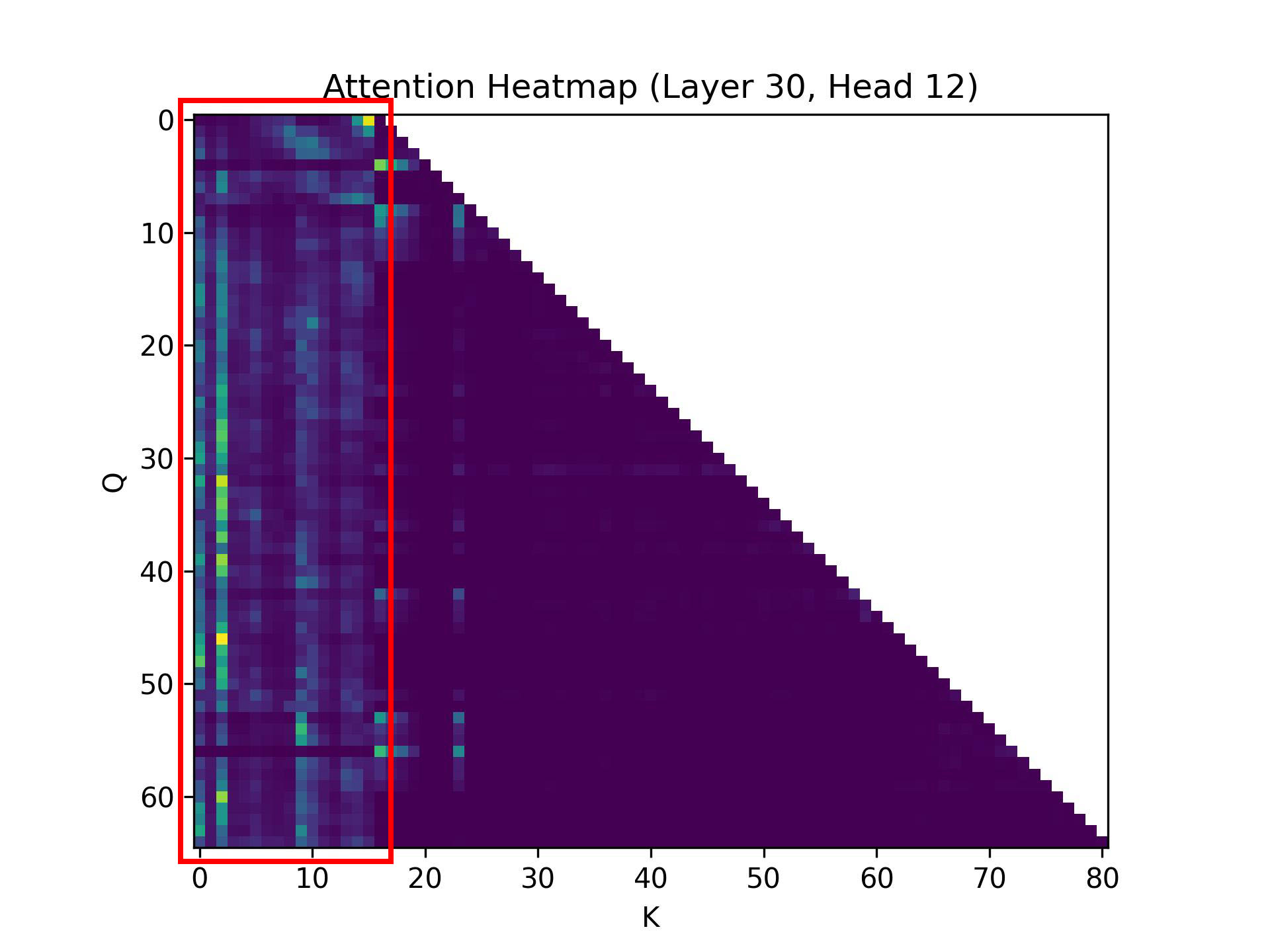} %
        \label{fig:image2}
    \end{minipage}\hfill
    \begin{minipage}[b]{0.28\textwidth}
        \centering
        \includegraphics[width=\textwidth]{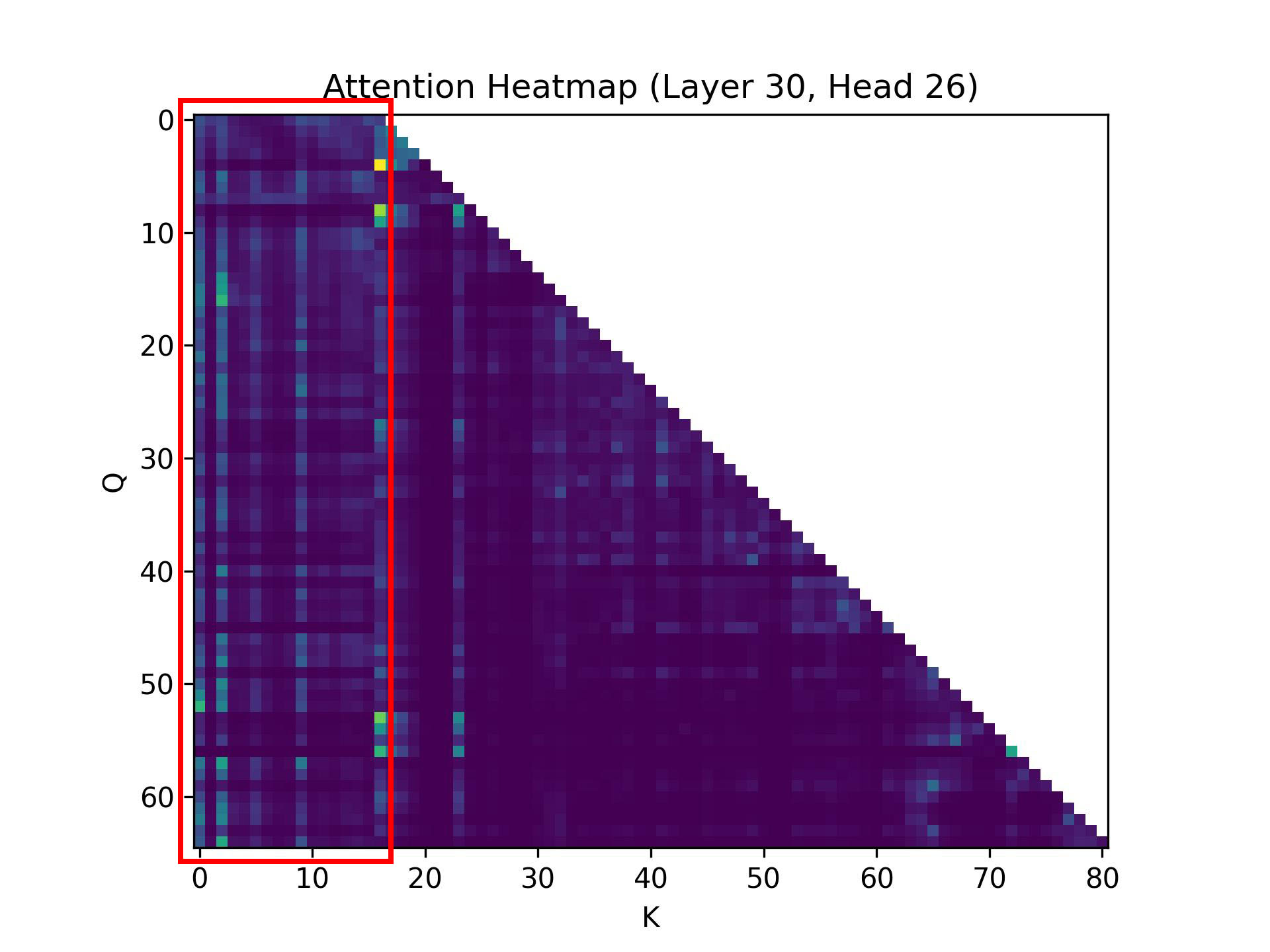} %
        \label{fig:image2}
    \end{minipage}\hfill
    \begin{minipage}[b]{0.28\textwidth}
        \centering
        \includegraphics[width=\textwidth]{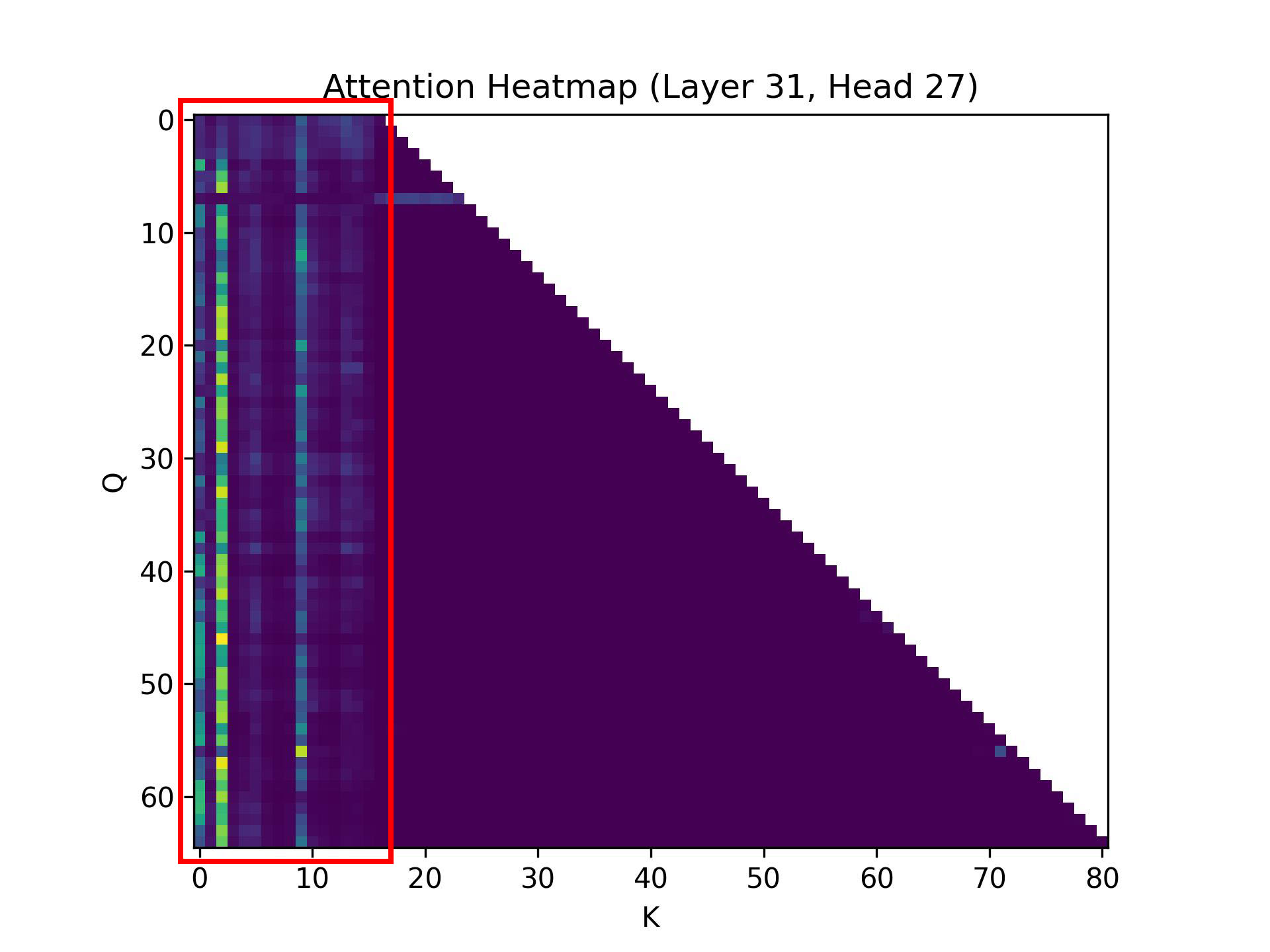} %
        \label{fig:image2}
    \end{minipage}\hfill
    \caption{Attention heamap of NarrativeQA in LongBench. The first 15 columns, marked by red rectangle lines, represent the attention weights corresponding to 15 candidate tokens. Since the final answer may rely on information from multiple chunks, we observe that many candidate tokens are assigned high attention weights. This suggests that FocusLLM effectively aggregates information from multiple chunks.}
    \label{Attention heamap for NarrativeQA in LongBench}
\end{figure*}

We visualized the information carried by candidate tokens when their Key/Value representations are concatenated with the tokens in the local context, and select a few representative heads in Figure \ref{Attention Heamap for Passkey Retrieval task} and Figure \ref{Attention heamap for NarrativeQA in LongBench}. We found that different patterns emerge in Passkey Retrieval and NarrativeQA tasks. The y-axis corresponds to the query representations of tokens in the local context, and the x-axis corresponds to the key representations of candidate tokens combined with the local context tokens. Therefore, the first few columns of the heatmap represent the contribution of candidate tokens to the local context. We made the following interesting observations: \textbf{i)} Not all heads in all layers attend to candidate tokens, and higher layers attend to candidate tokens more frequently than lower layers. This is likely because higher layers are more critical for the final representation.
\textbf{ii)} In Passkey Retrieval task, only one chunk contains passkey information, while the others are noises. As a result, we observe that a single candidate token receives high attention (a single column is highlighted), while other candidate tokens are ignored.
\textbf{iii)} In NarrativeQA task, the final answer may rely on information from multiple chunks, so we see that many candidate tokens are assigned higher attention weights. In summary, the result indicates that FocusLLM effectively ignores noise and aggregates information from multiple chunks.

\end{document}